\documentclass[10pt,twocolumn]{article}

\usepackage[top=2.5cm,bottom=2.5cm,left=2cm,right=2cm]{geometry}
\usepackage[utf8]{inputenc}
\usepackage[T1]{fontenc}
\usepackage{lmodern}
\usepackage[expansion=false]{microtype}
\usepackage{booktabs}
\usepackage{float}
\usepackage{array}
\newcolumntype{L}[1]{>{\raggedright\arraybackslash}p{#1}}  
\usepackage{graphicx}
\usepackage{amsmath,amssymb}
\usepackage{xcolor}
\usepackage{caption}
\usepackage{titlesec}
\usepackage{url}
\usepackage{needspace}
\usepackage{changepage}
\usepackage{balance}
\usepackage[numbers,sort&compress]{natbib}
\definecolor{linkblue}{HTML}{2E6DAD}
\definecolor{rulegrey}{HTML}{555555}
\usepackage[breaklinks,colorlinks=true,urlcolor=linkblue,
            citecolor=black,linkcolor=black]{hyperref}
\usepackage{xurl}

\graphicspath{{figs/}}
\newcommand{\headingset}{\raggedright\hyphenpenalty=10000\exhyphenpenalty=10000}
\titleformat{\section}{\normalfont\large\bfseries\headingset}{\thesection.}{0.6em}{}
\titleformat{\subsection}{\normalfont\normalsize\bfseries\headingset}{\thesubsection.}{0.6em}{}
\titlespacing*{\section}{0pt}{1.4ex plus .3ex}{0.7ex}
\begin{document}

\twocolumn[{%
\begin{center}
  \raisebox{-0.15\height}{\includegraphics[height=9mm]{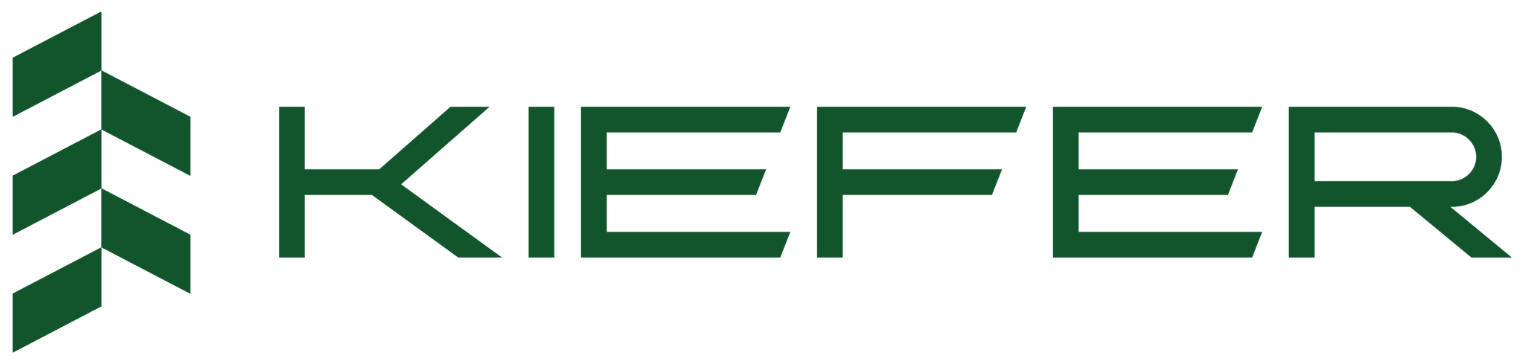}}%
  \hspace{15mm}%
  \raisebox{-0.15\height}{\includegraphics[height=10mm]{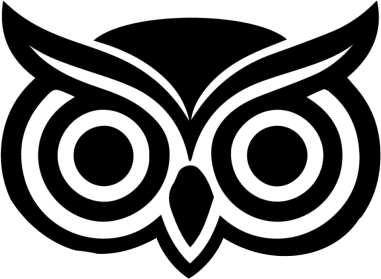}}\\[3.5mm]
  {\fontsize{19}{22}\selectfont\bfseries Measuring Language Transfer in Robot Policies}\\[1.8mm]
  {\fontsize{13}{15.5}\selectfont\itshape Adding Greek to a Cosmos3 vision-language-action policy}\\[3mm]
  {\fontsize{11}{13}\selectfont\bfseries Ayoub Kirouane\textsuperscript{1}
   \hspace{2em} Georgios Giaples\textsuperscript{1}
   \hspace{2em} Christos Petrocheilos\textsuperscript{1}}\\[1.5mm]
  {\fontsize{9.5}{11.5}\selectfont
   \textsuperscript{1}Sophea AI, KIEFER SA, Athens, Greece\\
   \{a.kirouane, g.giaples, c.petrocheilos\}@kiefer.gr\\
   models@sophea.ai}\\[1mm]
  {\fontsize{9}{11}\selectfont\color{rulegrey} September 2026}
\end{center}
\vspace{3mm}
\begin{minipage}{\textwidth}
\begin{adjustwidth}{3em}{3em}
\begin{center}{\fontsize{10.5}{12.5}\selectfont\bfseries Abstract}\end{center}
\vspace{1mm}
\small\noindent
Robot foundation models are trained and evaluated almost entirely in English,
and no robot demonstration corpus exists for most languages. We ask what it takes
to add one, using Greek and an open vision-language-action stack whose Greek is machine
rephrased from English under a mandated glossary, none of it human-authored. That part is
easy. The hard part is knowing whether it worked, and this paper is mostly about
that. Five instruments that a practitioner would reach for first all report
success where there is none: a colour-histogram coherence metric doubled twice on
generations that were pure noise; a standard single-goal benchmark scored $84.6\%$
under Greek instructions and $82.6\%$ under deliberately \emph{wrong} ones; a ten-goal
suite credited a policy with Greek instruction-following that a ninety-task suite shows
to be marginal at best, under three points over its own control on every seed;
training loss ranks six policies within $1.4\%$ of each other while their Greek
ability spans a factor of $7.2$; and single-run comparisons between recipes are
uninterpretable, because target-language success moves $31.6$ points on the random
seed while English moves $1.0$. Five of our own conclusions did not survive
contact with these controls; we report each retraction with the evidence that
forced it, because a claim that survives a control it could have failed is worth
more than one that was never tested. What survives is a small, checkable core: demonstrations in the
target language are \emph{necessary} but not \emph{sufficient}. A multilingual tower
transfers nothing to the action pathway without them (English-only training leaves Greek
at its own wrong-instruction floor, across three seeds), and target-language
demonstrations on their own are little better: on ninety tasks, across three seeds, a
Greek-only policy's margin over its own wrong-instruction floor never exceeds $2.7$ points
while a bilingual policy's never falls below $6.7$. A bilingual policy follows Greek on roughly half the episodes English
reaches on ten goals and two fifths of them on ninety tasks, with no architecture change;
much of the ten-goal number is that pipeline's glossary-bound phrasing rather than the
language, a penalty that training on seven phrasings per task halves; and warm-starting from a
language-adapted world model (one run) or unfreezing the text tower (three seeds)
both make things worse.
We also translated a real-robot corpus an order of magnitude larger in episodes
than every simulated suite we had combined, and explain precisely why we cannot
score it. Our
recommendations are technical and cheap: build the null before the metric, and
replicate before believing.
\end{adjustwidth}
\vspace{4mm}
\end{minipage}
}]

\section{Introduction}

Robot foundation models are built, trained, and evaluated in English. Their
demonstration corpora are English-annotated, their benchmarks issue English
instructions, and the language towers they inherit from vision-language
pretraining are strongest in English. For a speaker of any other language this
is a hard wall: there is no Greek robot demonstration dataset, and collecting one
means teleoperating a robot for months in order to reproduce capabilities that
already exist behind an English interface.

This paper asks what the cheap path buys. We translate an existing corpus by
machine, change nothing about the architecture, and measure what transfers. The
translation took hours and worked. Measuring it took the rest of the study, and
that asymmetry is our main finding: for a claim of the form ``the policy follows
language~$L$'', the bottleneck is not the data or the model but the instrument.

We therefore report this as a study rather than a system. Its spine is a set of
controls and what they destroyed. Every positive claim below is paired with a
condition under which it would have failed, and where the control won, we say so:
five conclusions we had drawn and written up did not survive replication or a
null, and they appear here as retractions with the evidence that forced them
rather than as omissions. We think this is the more useful artifact. The recipe
we ended with is small enough to state in a sentence, and the reader who copies it
without copying the controls will not know whether it worked for them.

What survives is layered, and the layers are the contribution: the choice of base
model decides whether the project is possible at all; a multilingual tower is
necessary but transfers nothing by itself; demonstrations in the target language
are what convert tower knowledge into behaviour; and what the target language
actually learns is largely the training translator's phrasing, a penalty that
training on several phrasings per task halves on the ten-goal suite. Two interventions that seem
obviously helpful (initializing from a world model adapted to the target language,
and unfreezing the tower so it can adapt) make things worse, and we report them as
such.

Measuring any of this turns out to require care, because the benchmarks that
would normally answer the question cannot. Where each scene admits exactly one
trained goal, success rates are insensitive to the instruction: prior work
finds that vision-language-action policies largely ignore language on such
suites~\cite{liberoplus2025}, and our own policies reproduce that pathology
(84.6\% under Greek instructions, 82.6\% under deliberately wrong ones). Working
cross-lingually gives us an unusually clean instrument for this. A language the
policy provably cannot follow (established by an identical policy trained
without that language, which performs at chance) provides a guaranteed null,
with a null guaranteed by construction rather than assumed, at the cost of training a
second policy to establish it.

\paragraph{Contributions.}
\begin{enumerate}
  \item \textbf{The tower is the bottleneck.} Supervised fine-tuning cannot teach
  Greek grounding to a video world model whose text tower was pretrained
  (effectively) on English: across a dose ladder from frozen tower to full
  learning-rate tower training at triple duration, Greek-conditioned generation
  remains noise. Swapping to a base model with a multilingual tower, the same
  stock recipe yields coherent Greek-conditioned scenes; the judged result below
  comes from a subsequent $6{,}836$-clip run (70/30 Greek/English) built on those
  $1{,}222$ captions plus LIBERO renders.
  \item \textbf{Grounding transfers by level (single judge, validated for
  coherence only).} The localized world model produces
  coherent in-domain scenes for 64\% of Greek prompts, where the base model
  produced none that we judged coherent on inspection, but matches the specific
  prompt content in 0\% of cases, reaching a partial match on 30\%
  (English: 100\% coherent, 90\% good). Greek conditioning transmits domain
  reliably and specifics only partially.
  \item \textbf{Negative transfer from world model to policy (one run per arm).} Warm-starting the
  policy from the Greek-adapted world model, the intuitive two-stage pipeline,
  \emph{degrades} both languages (English 79.2\% vs.\ 96.4\% from base; Greek
  13.6\% vs.\ 48.6\%). Video-generation fine-tuning appears to erode
  action-relevant representations of the base model.
  \item \textbf{A working, honestly-scoped Greek policy, and its mechanism.}
  Bilingual demonstration sampling yields 48.6\% Greek task success on ten goals,
  and 27.4\% (three-seed mean) on ninety tasks, from machine-produced Greek alone. Per goal this is bimodal rather than uniform,
  and the wrong-instruction residue is command-dependent rather than a
  fixed fallback, though we could not identify the feature that governs it
  (Section~\ref{sec:study2}).
  \item \textbf{A cross-lingual attribution control.} Prior work shows that
  vision-language-action policies largely ignore language on standard
  suites~\cite{liberoplus2025}; we reproduce it in a new regime (84.6\% Greek versus 82.6\% wrong
  instructions on a single-goal suite) and contribute a variant whose null is
  \emph{guaranteed by construction}: an identically trained policy that provably cannot
  follow the probe language. This complements perturbation-based controls rather than
  replacing them; it buys a null we do not have to assume, at the cost of needing a second
  trained policy.
  \item \textbf{Replication at scale, three seeds per arm.} On a ninety-task suite
  (uniform chance $1.1\%$, but $22\%$ for a policy that reads the scene and ignores the
  instruction) the bilingual policy's margin over its own wrong-instruction
  control is $6.7$--$7.1$ points on every seed, while a policy trained on Greek
  demonstrations alone reaches at most $2.7$; every bilingual seed exceeds every
  Greek-only seed. The seed instability that dominates the ten-goal results shrinks
  from $31.6$ to $2.4$ points, so it was largely an artifact of ten clusters.
\end{enumerate}

\section{Related Work}
\label{sec:related}

\paragraph{Benchmarks passable without the modality they advertise.} That a
multimodal agent can score well without using one of its modalities is an old
and repeatedly rediscovered result. Balanced VQA was built because image-blind
question priors solved the original benchmark~\cite{goyal2017vqa2,agrawal2018vqacp};
unimodal ablations matched full models in vision-and-language
navigation~\cite{thomason2019shifting}; prompt-based classifiers perform nearly as
well with irrelevant prompts~\cite{webson2022prompt}. In imitation learning the
mechanism has a name: a policy latches onto whichever observed variable best
predicts the demonstrated action, which is causal confusion~\cite{dehaan2019causal},
and scene identity is a far better predictor than instruction text whenever a
scene determines its goal. Benchmark design has responded: CALVIN chains
instructions in a shared scene so affordance cannot substitute for
language~\cite{mees2022calvin}, and the LIBERO suite family varies goal
independently of scene~\cite{liu2023libero}.

Most directly, prior work established our methodological conclusion in English.
LIBERO-Plus perturbs seven factors across VLA models and reports that they are
``largely insensitive to language variations'' and ``tend to ignore language
instructions completely''~\cite{liberoplus2025}. We therefore do not claim the
observation as novel. Our contribution on this axis is a complementary instrument and a confirmation in a
different regime: rather than deleting or paraphrasing an instruction, we issue it in a
language the policy is \emph{independently proven} not to follow (an otherwise identical
policy trained without target-language demonstrations scores at chance). The null is then
guaranteed by construction rather than assumed, which removes the question of how much a
paraphrase ought to matter, at the cost of a second trained policy. It does not supersede
perturbation designs, which need no such policy and cover factors a language swap cannot.

\paragraph{Vision-language-action models and language conditioning.} Success
rates on LIBERO and similar suites are routinely reported as evidence of
instruction following~\cite{brohan2023rt2,kim2024openvla,octo2024,black2024pi0},
building on language-conditioned imitation learning~\cite{lynch2021mcil}. Our
policies use the Cosmos platform~\cite{nvidia2025cosmos}; the localization
question we ask is orthogonal to model scale. Our own search did not surface prior work that trains and evaluates a
vision-language-action policy in a language other than English, but this is a fast-moving
area and we make no priority claim: we expect concurrent work and would welcome
correction. We frame the contribution as instruments rather than as a result that beats a
baseline.

\paragraph{Multilingual transfer and machine-translated localization.} Zero-shot
cross-lingual transfer from multilingual encoders is well
characterized~\cite{conneau2020xlmr,hu2020xtreme}, and in instruction tuning a
small multilingual admixture over a strong high-resource base outperforms
monolingual target-language tuning~\cite{shaham2024pinch}. We set out to test
the embodied analogue of that result and can report only that our data point the
same way and, on the larger suite, resolve it in paired form: on ten goals bilingual
training beats Greek-only on average across three seeds per arm but not separably; on
ninety tasks, again at three seeds per arm, every bilingual seed's margin over its own
control exceeds every Greek-only seed's (Section~\ref{sec:study2}). What we can state
cleanly in either case is that without demonstrations in the target language a policy
does not follow that language at all, and that with them alone it barely does. Multilingual grounding has been studied in
navigation~\cite{ku2020rxr} but not, as far as we know, in manipulation. Because
our entire target-language corpus is machine generated, machine artifacts are a live
confound~\cite{artetxe2020translation}, though the pipeline rephrases rather than
translates; we return to this in Section~\ref{sec:limits}.

\paragraph{What fine-tuning costs a pretrained representation.} Full fine-tuning
distorts pretrained features and degrades out-of-distribution
performance~\cite{kumar2022finetuning,wortsman2022wiseft}, a specific case of
catastrophic forgetting~\cite{kirkpatrick2017ewc}. For VLAs specifically,
prior work argues that action-training gradients degrade the backbone's
semantic knowledge and proposes insulating the backbone from
them~\cite{driess2025ki}. Our tower-unfreezing result is consistent with that
prediction and extends it to the cross-lingual case, where the pretrained
representation is the \emph{only} source of target-language competence and its
degradation is therefore unusually visible.

\paragraph{World models as policy initializers.} Video-generative pretraining
has been reported to help manipulation policies~\cite{wu2024gr1,du2023unipi},
and generated trajectories with inverse-dynamics labels are an established route
to policy data~\cite{baker2022vpt,dreamgen2025}. Our negative-transfer result
sits in tension with that literature: warm-starting from a video-generation
checkpoint degraded our policies in both languages. We do not claim to overturn
those results, whose setups differ from ours in objective, scale, and
architecture; we report the discrepancy and the one-variable comparison that
produced it.

\section{Setup}
\label{sec:setup}

\begin{figure}[t]
\centering
\includegraphics[width=\columnwidth]{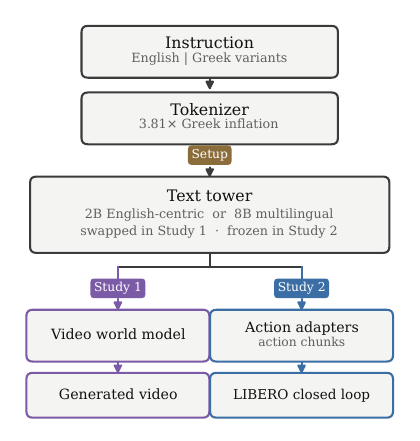}
\caption{The Cosmos3 stack as this paper touches it. An instruction enters as
pipe-separated English and Greek variants, is segmented by a tokenizer that inflates Greek
by $3.81\times$ (Figure~\ref{fig:tokenizer}), and conditions a text tower shared by both
branches. Section~\ref{sec:study1} swaps the tower and reads the video branch;
Section~\ref{sec:study2} freezes it, trains only the action adapters, and reads the
closed-loop branch. No block's internals are drawn, because none are varied here.}
\label{fig:arch}
\end{figure}

\paragraph{Model stack.} We use the Cosmos3 open robot-learning stack (Figure~\ref{fig:arch}): a video
world model in two variants, one with an English-centric 2B text tower
(``English-tower model'') and one whose tower is a multilingual 8B
vision-language model (``multilingual-tower model''), and an action policy
architecture that conditions on instructions through the same tower and decodes
action chunks through dedicated adapter modules. Policies are trained by
imitation on demonstration datasets in LeRobot format; evaluation is closed-loop
in the LIBERO simulator, with binary task success judged by the simulator's goal
predicates.

\begin{figure}[t]
\centering
\includegraphics[width=\columnwidth]{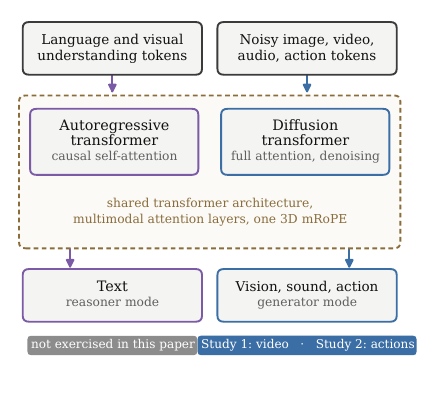}
\caption{The Cosmos3 backbone, redrawn from the vendor's description. An autoregressive
transformer for reasoning and a diffusion transformer for generation sit inside one
Mixture-of-Transformers: they share the transformer architecture, the multimodal attention
layers and a single 3D rotary position embedding, and differ in how they attend, causally
for reasoning and fully for denoising. Both of our studies exercise the generator surface,
Study~1 through generated video and Study~2 through actions; the reasoner surface is never
used in this paper. No layer count, width or routing rule is drawn, because the vendor
states none and we measure none.}
\label{fig:mot}
\end{figure}

\paragraph{Why the stack looks like this.} Cosmos3 is an omnimodal world model built on a
unified Mixture-of-Transformers architecture: an autoregressive transformer for reasoning
and a diffusion transformer for generation share one set of multimodal attention layers and
a single 3D rotary position embedding over space and time, so language, vision, audio and
action are handled as modalities of one model rather than by separate encoders joined at the
output (Figure~\ref{fig:mot})~\cite{nvidia2026cosmos3}. In reasoner mode, text and visual tokens run through causal
self-attention; in generator mode, noisy image, video, audio and action tokens are denoised
under full attention. This matters for what follows in one specific way: the instruction
pathway a policy conditions on is the same pathway the world model reads, which is why
Section~\ref{sec:study1} can interrogate the tower through generated video and
Section~\ref{sec:study2} can interrogate it through actions, and why a result about one is
evidence about the other. We take this description from the vendor's documentation and do
not verify it; nothing we measure opens the backbone or localizes where inside it a language
is represented.

\paragraph{Greek data, produced by machine.} No Greek robot data exists and we authored
none by hand. An LLM pipeline rephrases English into Greek imperatives ``as if originally
authored in Greek'' under a mandatory glossary, with anti-calque rules and a review pass: (a) 1{,}273
rich structured scene captions of a BridgeData~V2 subset~\cite{walke2023bridgedata},
and (b) all 53{,}207 unique task instructions of the two suites: 53{,}096 from
DROID~\cite{khazatsky2024droid}, which carries them across 57{,}639 success
episodes, and 111 from LIBERO~\cite{liu2023libero}. Quality was
audited by Greek-character ratio (mean $\geq 0.995$), structure preservation,
and spot review. The translated instruction set is dominated by DROID; what we can
do with it is limited by measurement rather than by data, and
Section~\ref{sec:droid} states that limit precisely. Bilingual training requires no loader changes: instruction
fields hold pipe-separated variants from which the loader samples uniformly at
each step, so appending the Greek translation to the English variants yields
50/50 language sampling for free.

\begin{figure}[t]
\centering
\includegraphics[width=\columnwidth]{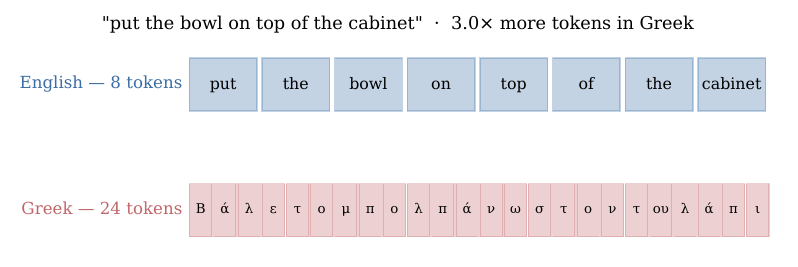}
\caption{One command as the multilingual tower's tokenizer sees it, the same
command quoted in Section~\ref{sec:setup}. English splits into eight word-like
pieces; the Greek translation splits into twenty-four, most of them single
letters. Across the ten evaluation instructions the inflation is
$3.81\times$ ($72$ tokens against $274$). This is a property of the substrate, not
of our method, and it remains entangled with representational depth as an
explanation of the residual gap. Greek script is shown here because the figure is
about sub-word segmentation, which transliteration would misrepresent.}
\label{fig:tokenizer}
\end{figure}

\paragraph{Tokenization.} The multilingual tower's tokenizer is far less
efficient in Greek than in English. Across the ten evaluation instructions it
produces 274 Greek tokens against 72 English ones, an inflation of $3.81\times$
(Figure~\ref{fig:tokenizer}),
and the segmentation is close to character-level: \emph{put the bowl on top of
the cabinet} is eight word-like tokens, while its Greek translation is
twenty-four pieces, most of them single letters. We report this here because it
is a candidate explanation for results below, and because it is a property of
the substrate rather than of our method.

\paragraph{Training configuration.} Both studies use the stock Cosmos3 SFT trainer on
$8\times$B200. Policies train for $2{,}000$ iterations from the base action-policy
checkpoint, AdamW at learning rate $5\times10^{-5}$, weight decay $0.05$, $500$ warm-up
steps on a cosine schedule, batch size one per device, text tower frozen, action chunks
of sixteen at 20\,FPS with a ten-dimensional frame-wise-relative action space and 6-D
rotations, agentview and wrist cameras at $256^2$. The multilingual tower is Qwen3-VL-8B-Instruct; the English-centric tower is the 2B
model of the same family. The instruction and caption translations were produced by
kimi-k3; the independent Greek renderings and the English rewordings by
gemini-3.7-flash and kimi-k3 respectively; and the coherence and caption-match judgements
of Study~1 by gemini-3.7-flash. We name these because a reader reproducing the study needs
to know that the ``independent'' translator and the judge are themselves particular models,
and because two of them recur in more than one role. World-model runs use the same trainer for $2{,}000$ iterations with the tower
frozen and only the generation pathway and its projections in the optimizer. Seeds are
$42$, $43$ and $44$ wherever three are reported. Configuration files accompany the paper.

\paragraph{Evaluation protocol.} Every policy is evaluated three ways on the
same tasks, seeds, and initial states: correct English instructions, correct
Greek instructions, and \emph{wrong} instructions (each task commanded with a
different task's Greek instruction). All headline numbers use 50 trials per task
(500 episodes per condition); exploratory readings at 10 trials per task are
noted where relevant and differed from the replication by at most 3.4 points. For world-model outputs we report judged coherence and caption match
(a vision-language model judged each clip against the English source caption
regardless of generation language). We flag a weakness in this instrument: its
calibration set labelled Greek-prompted clips as noise by assumption, which is
part of what it is used to measure, and it validated only the binary
coherence judgement, not the three-way content-match judgement we report
in Section~\ref{sec:study1}. We use it after finding that a
color-histogram similarity metric was fooled twice by noise whose palette
drifted toward the reference (Section~\ref{sec:lessons}).

\paragraph{Statistical treatment.}
\label{sec:stats}
Each evaluation condition comprises ten goals $\times$ 50 trials. Success is
strongly clustered by goal (per-goal rates span the full range within a single
condition), so episodes are not independent draws and a binomial interval over
500 episodes badly understates uncertainty. We therefore treat the goal as the
sampling unit and report a paired task-level bootstrap (20{,}000 resamples of the
ten goals, applied to both arms of a comparison simultaneously). With only ten clusters the percentile
bootstrap is mildly \emph{anti}-conservative (nominal 95\% intervals cover
roughly 89--91\% under our own per-goal rate profiles), so we corroborate every
conclusion that matters with an exact paired sign-flip permutation test, whose
smallest attainable two-sided $p$ at $n=10$ is $0.002$. Between-seed variance is not negligible and we have
measured part of it. Retraining the reference policy with only the seed changed
moves Greek success across 48.6/80.2/55.2\% at three seeds, a range of 31.6
points, while English moves 1.0 (96.4/95.6/96.6\%) and the wrong-instruction
floor stays low (4.4/2.2/0.8\%). On the ninety-task suite the same three-seed
comparison moves Greek by 2.4 points, so much of that spread belongs to the
ten-cluster suite rather than to the recipe (Section~\ref{sec:study2}). The
high-resource language is stable under reseeding; the low-resource
language is not. That is itself a result, and a caution for anyone reporting
single-run low-resource numbers. Its consequence here is that our
\emph{cross-policy} comparisons cannot be read at the precision their point
estimates suggest. We have since replicated four arms at three seeds each, which
is what allows the contrasts in Section~\ref{sec:study2} to be stated at all;
arms still at one run are marked where they appear. Even at three seeds, a difference of
thirty points between two singly-trained policies is not clearly larger than the
spread we observe between seeds of one policy. We mark those claims as
directional throughout, and we replicated the arms that carry practical advice.
Comparisons \emph{within} a policy (the wrong-instruction conditions, the
phrasing ladder, the per-goal profile) share a checkpoint, so the additive seed
effect cancels; what does not cancel is any seed-by-condition interaction, and we
measure one (the translator penalty itself spans $16.4$ points across seeds of the
single-phrasing arm). They are far less affected, not unaffected. Counting
every inferential comparison in the text, not only those we tabulate, we make at
least twenty, and we apply no multiplicity correction; Appendix~\ref{app:comparisons}
lists them all with their sampling units so a reader can apply one. Only the within-policy
contrasts, whose task-level tests reach $p=0.002$, would survive one; no
run-level contrast can, because the smallest attainable $p$ at three runs per arm
is $0.100$. Small differences should be read as exploratory.

Our cross-policy claims therefore rest on a single decision rule, which we state
once here: every seed of one arm exceeds every seed of the other on the paired
within-policy quantity. At three runs per arm this is the exact permutation
test's floor, $p=0.100$, so it carries a one-in-ten false-positive rate per
comparison and we do not report it as significance. We use it as an ordering
statement, always alongside the run-level effect size (for the ninety-task
margins: a $4.0$-point gap between the smallest bilingual margin, $6.7$, and the
largest target-only one, $2.7$).

\section{Study 1: Localizing the World Model}
\label{sec:study1}

\noindent{\small\itshape Targets in Figure~\ref{fig:arch}: the text tower, swapped between its two
variants, and the video world model it conditions.}

\paragraph{An English tower cannot learn Greek by SFT.} We fine-tuned the
English-tower world model on 855 Greek and 367 English rich captions, the $1{,}222$ of the
$1{,}273$ that survived caption-quality filtering, under a dose
ladder: generation pathway only (tower frozen); tower at $0.1\times$ the base
learning rate; tower at the full rate for triple the iterations; and a
100\%-Greek continuation. In every configuration, Greek-conditioned generation
remained structureless noise while English-conditioned generation improved
markedly (judged frame inspection; the histogram metric incident of
Section~\ref{sec:lessons} occurred here). We conclude that moderate-scale caption
SFT cannot induce a new language in a tower that lacks it.

\begin{figure*}[t]
\centering
\includegraphics[width=0.86\textwidth]{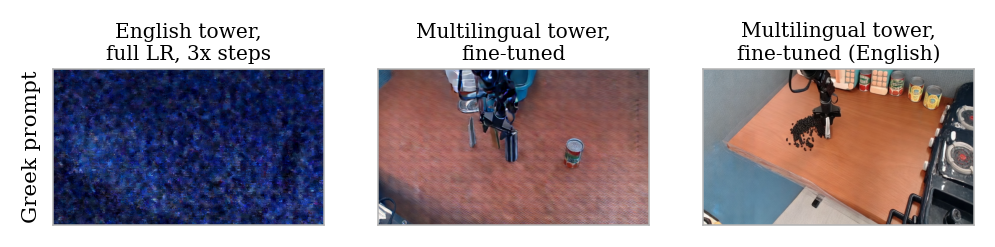}
\caption{Generation from the same held-out Greek caption (mid-clip frame). The
English-tower model produces no Greek-conditioned structure even at full
learning rate and triple duration (left); the multilingual-tower model, given the
identical recipe, produces a recognizable manipulation scene (centre). The
English-prompted reference (right) shows the content fidelity Greek does not
reach. Intermediate rungs of the dose ladder are omitted; all are
indistinguishable from the left panel.}
\label{fig:doseladder}
\end{figure*}

\paragraph{A multilingual tower localizes cheaply, but partially.} The
multilingual-tower base model also produces noise from Greek prompts, so tower
comprehension alone is insufficient: the generation pathway has never seen
Greek-derived conditioning. However, the \emph{stock} fine-tuning recipe (tower
frozen, generation pathway trained) on the same captions, followed by a larger
mixed run (6{,}836 clips, 70/30 Greek/English), produces genuinely coherent
Greek-conditioned robot scenes (Figure~\ref{fig:doseladder}).

\paragraph{Quantifying the transfer level.} Judged on 51 held-out scenes per
language, English generation is fully coherent with a good content match on 90\%
of clips (95\% Wilson interval $[79, 96]$; the Greek intervals below are
$[50, 76]$ for coherence and $[0, 7]$ for good match, so ``none'' is an upper bound
of seven percent rather than a demonstrated zero). Greek generation is coherent on 64\% of the 50 clips that returned an
evaluable judgement (of 51 submitted) but achieves a good content match on \emph{none}, and a
partial match on 30\%. Greek conditioning reliably selects the right \emph{kind}
of scene and not the described content: grounding transferred at the scene level,
not the content level.

\section{Study 2: Does the Policy Read?}
\label{sec:study2}

\noindent{\small\itshape Targets in Figure~\ref{fig:arch}: the text tower, held frozen, and the action
adapters, which are the only modules trained.}

\begin{figure*}[t]
\centering
\includegraphics[width=0.86\textwidth]{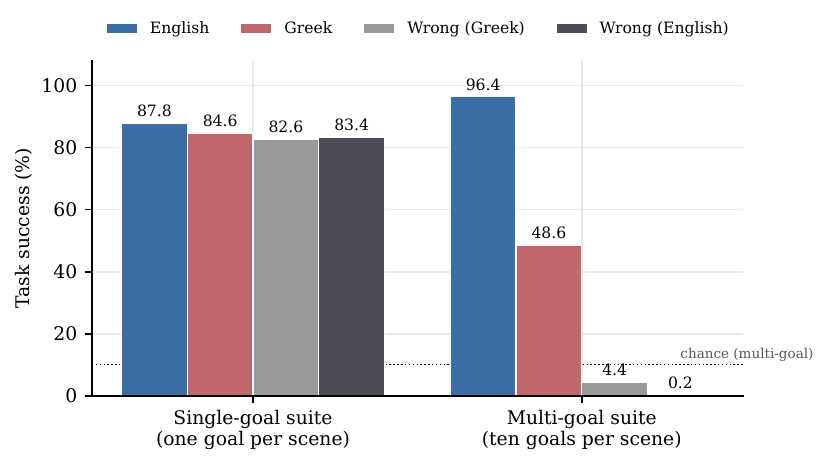}
\caption{Why single-goal benchmarks cannot test language. On a suite where each
scene admits one trained goal (left), success is nearly identical for correct
English, correct Greek, and deliberately wrong instructions. On a suite where ten
goals share a scene (right), the same evaluation separates them.}
\label{fig:attribution}
\end{figure*}

\paragraph{Single-goal suites cannot attribute language.} We trained a policy on
the LIBERO-10 suite with bilingual instruction sampling, warm-started from the
localized world model. It performs well: 87.8\% success under English
instructions and 84.6\% under Greek. Read naively, this is a bilingual robot.
The wrong-instruction control refutes that reading: commanded to perform a
\emph{different} task, the policy still succeeds 82.6\% of the time in Greek and
83.4\% in English (Table~\ref{tab:policy}; Figure~\ref{fig:attribution}). The English
column carries this argument. This policy is the warm-started recipe, which the
discriminative suite later shows reads Greek at chance ($13.6\%$), so its Greek
column cannot distinguish a suite that hides language from a policy that never had
it; its English column can, because the same policy reads English at $79.2\%$ on the
discriminative multi-goal suite, well clear of that suite's floor. All four conditions lie within six
points of one another. A task-level interval does exclude zero for a small
effect of the Greek instruction on this suite ($+2.0$ points over wrong
instructions, 95\% CI $[+0.6, +3.6]$, though the exact paired test gives
$p=0.055$) and not for English ($+5.2$, CI $[-1.2, +12.4]$), so we do not claim
the instruction is entirely unused; we claim it is nearly irrelevant, and
negligible beside the 44.2-point separation the multi-goal suite produces. Each LIBERO-10 scene admits one trained goal, so the
policy executes what the scene affords and the instruction is ignored. No
language claim survives on such a suite without this control.

\paragraph{A discriminative protocol.} The LIBERO-Goal suite places ten goals in
a single scene, making the instruction the only goal-selection signal. Here the
three-way evaluation becomes discriminative, and it immediately falsified two of
our own assumptions.

\paragraph{Negative transfer from the world model.} The policy warm-started from
the Greek-adapted world model reaches 79.2\% English but only 13.6\% Greek,
statistically at the 10\% chance level (and 9.4\% under wrong instructions).
Retraining the \emph{same} recipe from the base model, discarding the world-model
warm start, improves both languages dramatically: 96.4\% English and 48.6\% Greek
(wrong: 4.4\%). The intuitive transfer pipeline was not merely unnecessary but
harmful. We offer video-generation fine-tuning eroding action-relevant
representations as one explanation, but a second is not excluded by this design:
the world model was adapted on BridgeData~V2, a real WidowX corpus, while the policy
is trained and evaluated on a simulated Franka in LIBERO, so the warm start also
carries a domain shift. Separating the two needs a world model adapted in-domain,
which we did not run. The world model remains valuable as a
synthetic-data engine, but its weights should not seed the policy.

\begin{table*}[t]
\centering
\caption{Closed-loop task success, 50 trials per task (500 episodes per cell).
The wrong-instruction column commands each task with a different task's
Greek instruction; the italicised rows repeat that condition in English.}
\label{tab:policy}
\begin{tabular}{llccc}
\toprule
Suite & Policy & English & Greek & Wrong \\
\midrule
LIBERO-10 (1 goal/scene) & bilingual, warm-started & 87.8\% & 84.6\% & \textbf{82.6\%} \\
\multicolumn{2}{r}{\emph{\quad same policy, wrong instruction in English}} & & & \textbf{83.4\%} \\
LIBERO-Goal (10 goals/scene) & bilingual, warm-started & 79.2\% & 13.6\% & 9.4\% \\
LIBERO-Goal (10 goals/scene) & bilingual, from base & \textbf{96.4\%} & \textbf{48.6\%}$^\dagger$ & 4.4\% \\
\multicolumn{2}{r}{\emph{\quad same policy, wrong instruction in English}} & & & 0.2\% \\
\midrule
LIBERO-Goal & English-only, from base & 98.2\% & 9.0\%$^\dagger$ & 7.6\% \\
LIBERO-Goal & Greek-only, from base & 10.0\% & 20.4\%$^\dagger$ & 2.2\% \\
LIBERO-Goal & bilingual, tower unfrozen & 82.4\% & 18.0\%$^\dagger$ & 4.8\% \\
LIBERO-Goal & bilingual, 7 Greek phrasings/task & 92.8\% & 52.0\%$^\dagger$ & 3.0\% \\
\bottomrule
\end{tabular}

\vspace{2pt}
{\footnotesize $^\dagger$Single run. We later replicated these four rows at three
seeds each; Greek is highly seed-dependent while English is not. Greek across seeds:
bilingual $48.6/80.2/55.2$, English-only $9.0/9.2/9.2$, Greek-only $20.4/66.6/23.8$,
tower-unfrozen $18.0/13.4/29.8$, seven-phrasings $52.0/64.8/84.8$. \textbf{Rows in this
table should not be compared against one another on their Greek column}: a 31.6-point
seed range swamps most of the differences. Only the English-only row is separable from
the baseline, and it lands on its own wrong-instruction floor. See
Section~\ref{sec:stats}.}
\end{table*}

\paragraph{What bilingual sampling buys, and what the mean hides.} The from-base
bilingual policy succeeds on 48.6\% of Greek-commanded episodes (task-level
bootstrap 95\% CI $[25.0, 72.2]$; ten goals is ten clusters, and the interval is
correspondingly wide), with zero architecture changes and purely
machine-translated instructions. The mean is a
poor description of the behaviour. Per goal, Greek success is sharply bimodal,
with five goals at $0.72$--$0.98$ and five at $0.00$--$0.24$ and nothing in
between, while English is at least $0.88$ on all ten (Table~\ref{tab:pertask}). The split
reproduces in an independent smaller run at the same seed. It does not survive
reseeding: profiling all three seeds gives five, nine and six of ten goals above $50\%$,
and no goal fails in all three, so the split is a property of the run and not of the
recipe (Conclusion).

Two observations bear on the mechanism. First, the instruction set contains a
near-minimal pair: \emph{put the bowl on top of the cabinet} and \emph{put the
wine bottle on top of the cabinet} differ in Greek only in the object noun
phrase, and score $0.94$ and $0.00$ respectively. Both content words are
independently grounded elsewhere, so the failure is not vocabulary but
discrimination within a shared referent. Second, the wrong-instruction condition
leaves a small, structured residue rather than a uniform floor, and that residue
is \emph{command-dependent}. Re-running it with every distractor reassigned (a
different rotation of the same instruction set) moves the residue entirely: the
single goal carrying all of it under the first assignment drops to zero, and five
different goals become non-zero. A policy falling back on one dominant behaviour
whenever Greek is uninformative would produce the same cell under both
assignments; it does not. The policy is therefore processing the content of a
Greek instruction even when that instruction is wrong.

We tested one specific account of what it extracts (latching onto a shared head
noun), and the data do not support it: of the six goals whose reassigned
distractor shares a noun stem with them, only three produce any success, and two
of the five non-zero cells share no stem at all. We therefore report the
structure without claiming its cause. Greek conditioning selects behaviour that
depends on the instruction's content, discriminates poorly between goals that
share a referent, and we could not identify the feature that decides which wrong
command elicits which behaviour.

A third observation reframes all of these numbers. Our English-only policy is a
measured per-goal null: it cannot follow Greek, yet its Greek success is not a
flat 10\% but $0.34$ on \emph{put the bowl on the plate}, $0.32$ on \emph{put
the bowl on the stove}, and near zero elsewhere. Read against that null rather
than against chance, the bilingual policy's profile changes materially: its
largest genuine gains are $+0.84$ on two goals, while on three goals (including
one the paragraph above would count as a Greek failure) it is at or below a
policy with no Greek at all. Pooled chance rates conceal this; per-goal nulls
should be reported wherever a guaranteed-null probe is available. The defensible
claim is therefore not 48.6\% Greek competence but \textbf{partial,
non-compositional referent-level grounding}, and a caution that aggregate
success rates conceal mechanism even after the wrong-instruction control has
been passed.

\begin{table*}[t]
\centering
\small
\caption{Per-goal success for the bilingual policy on the multi-goal suite
(50 trials per goal). English is at least 0.88 everywhere; Greek is bimodal, with no goal
between 0.24 and 0.72. The two goals set in bold differ in Greek only in the
object noun phrase. The single non-zero wrong-instruction cell arises where the
commanded instruction shares a head noun (\emph{cabinet}) with the goal.}
\label{tab:pertask}
\begin{tabular}{lccc}
\toprule
Goal & English & Greek & Wrong \\
\midrule
put the bowl on the plate & 1.00 & 0.98 & 0.00 \\
\textbf{put the bowl on top of the cabinet} & 0.96 & \textbf{0.94} & 0.44 \\
turn on the stove & 1.00 & 0.90 & 0.00 \\
put the wine bottle on the rack & 0.88 & 0.74 & 0.00 \\
push the plate to the front of the stove & 0.98 & 0.72 & 0.00 \\
\midrule
put the bowl on the stove & 0.98 & 0.24 & 0.00 \\
open the top drawer and put the bowl inside & 0.92 & 0.24 & 0.00 \\
open the middle drawer of the cabinet & 1.00 & 0.08 & 0.00 \\
put the cream cheese in the bowl & 0.94 & 0.02 & 0.00 \\
\textbf{put the wine bottle on top of the cabinet} & 0.98 & \textbf{0.00} & 0.00 \\
\bottomrule
\end{tabular}
\end{table*}

\paragraph{How much of this is Greek, and how much is our translator?} Every
Greek string the policy was trained on came from one machine-translation
pipeline, so its 48.6\% may reflect that pipeline's idiolect rather than the
language.

\begin{figure}[t]
\centering
\includegraphics[width=\columnwidth]{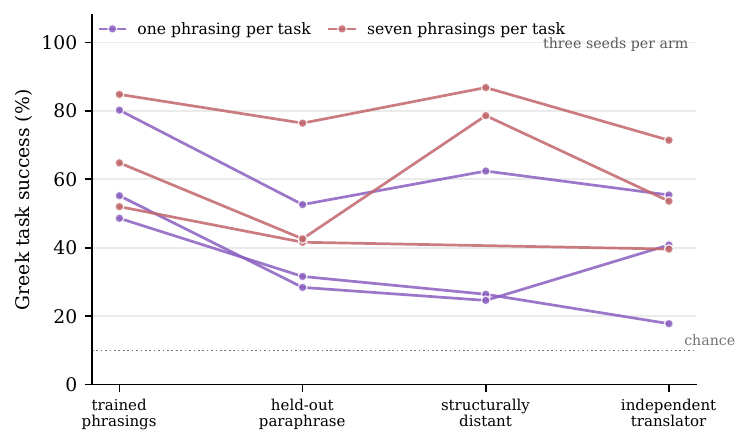}
\caption{Greek success against distance from the training translator, every run
drawn. Two things are visible and only one of them was in our first reading. The
diverse arm sits above the single-phrasing arm nearly everywhere, which
replicates. The descent is \emph{not} monotone: several runs score higher on the
structurally distant paraphrase than on the mild one, and one inverts by 36
points. A single-seed version of this figure showed a clean descent and we
believed it.}
\label{fig:ladder}
\end{figure}

We test this with three progressively more independent phrasings of
the same ten commands, at three seeds each. Held-out paraphrases from the same
generator score 31.6/52.6/28.4\% and 26.4/62.4/24.6\% --- ranges of 24 and 38
points, so the levels are seed artifacts and we quote them only to show their
spread. An independent translation (a different language model, not a native speaker, prompted to phrase
each command the way a Greek speaker would actually say it, sharing no string
with the training set) costs \textbf{23.3 points} on average: three seeds of this
recipe fall 48.6/80.2/55.2\% to 17.8/55.4/40.8\%, drops of 30.8, 24.8 and 14.4
points. We report the drop rather than the level deliberately. The level is not
stable across seeds (a 37.6-point range), whereas both of its terms come from the
same checkpoint, so the seed variance that dominates cross-policy comparisons here
cancels within a policy. What the degradation is \emph{not} is monotone in distance from the
training translator. An earlier single-seed reading of this ladder showed a clean
descent, and we took it as evidence of a gradient running outward from the
training translator's idiolect. Replicated, the ordering holds in two of five
paired comparisons and inverts by 36 points in one: the diversity policy at seed
43 scores 42.6\% on the mildly reworded set and 78.6\% on the structurally distant
one. Because both terms come from one checkpoint, seed variance cannot explain
this away. The honest statement is that changing the phrasing costs accuracy, and
that how much it costs is not predicted by how distant the phrasing looks to us
(Figure~\ref{fig:ladder}).

\begin{figure}[t]
\centering
\includegraphics[width=0.92\columnwidth]{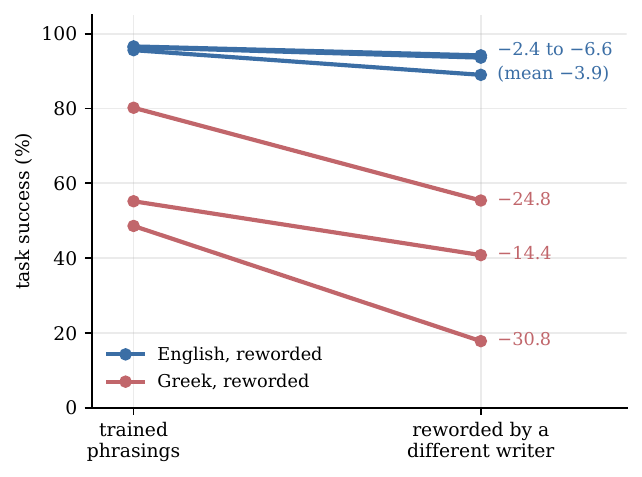}
\caption{The same rewording, in each language. Every line is one training run, moving from
the phrasings it trained on to the same commands reworded by a different writer; the label
is that run's drop. Rewording in English costs $2.4$ to $6.6$ points, rewording in Greek
$14.4$ to $30.8$. Every Greek drop exceeds every English drop, and both ends of each line
come from one checkpoint, so seed variance cancels within a line. This is the control that
separates translator idiolect from generic brittleness to unseen phrasings: if the policy
were merely overfitted to one phrasing per task, the blue lines would fall as steeply as
the red ones.}
\label{fig:enpara}
\end{figure}

\paragraph{Is it the translator, or is it one phrasing?} The drop above has an
alternative reading that the design so far does not exclude. The policy sees exactly one
\emph{English} phrasing per task as well, so it may simply be brittle to any rewording,
in which case the effect has nothing to do with translation. The control is cheap and we
ran it: the same ten commands rephrased by a different English speaker
(\emph{turn on the stove} becomes \emph{switch the stove on}), the same three
checkpoints, the same 50 trials per task. English success falls from $96.4/95.6/96.6\%$
to $93.6/89.0/94.2\%$, drops of $2.8$, $6.6$ and $2.4$ points, a mean of $3.9$. The Greek
drops on the same checkpoints are $30.8$, $24.8$ and $14.4$, a mean of $23.3$. Every Greek
drop exceeds every English drop, the smallest Greek being $14.4$ against the largest
English $6.6$, and both terms of each drop come from one checkpoint, so seed variance
cancels within a row. The policy handles unseen English phrasings nearly as well as
trained ones and loses roughly six times more when the rewording is in the low-resource
language (Figure~\ref{fig:enpara}).

\begin{figure*}[t]
\centering
\includegraphics[width=0.82\textwidth]{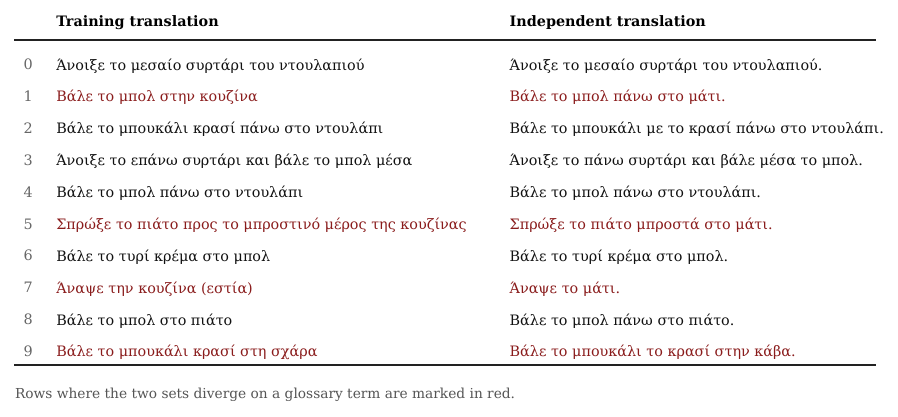}
\caption{The ten Greek instructions as the policy saw them in training, beside the
independently produced set used for the idiolect measurement. Goals 1, 5 and 7 show the
glossary at work: \emph{stove} becomes a word that also means \emph{kitchen}, where the
independent translator writes the word a Greek speaker uses for a hob ring. Goal 9's
independent rendering substitutes a word meaning \emph{wine store} for \emph{rack}, an
error in the independent set that we leave uncorrected, because correcting it after seeing
the scores would invalidate the measurement. Set as an image so the Greek renders in a
document whose text font has no Greek glyphs.}
\label{fig:strings}
\end{figure*}

That control is not sufficient on its own, and a reader should see why before believing it.
The two perturbations are not matched in strength. The English rewriter changed syntax and
left every content noun in place, none of ten; the independent Greek set changes a content
word in five of ten, including the term the training glossary had fixed
(Figure~\ref{fig:strings}). A perturbation that swaps the noun naming the target object is
larger than one that reorders a verb phrase, and our own analysis says the noun is what the
policy latches onto, so the comparison as it stands confounds language with perturbation
size.

We therefore ran a second English control matched on exactly that dimension. It substitutes
an everyday synonym for the object or surface in seven of the ten commands
(\emph{stove} to \emph{hob}, \emph{cabinet} to \emph{cupboard}, \emph{rack} to
\emph{shelf}), chosen so that the physical referent is unchanged and the substitute names no
other object in the scene. This is a \emph{more} aggressive noun perturbation than the Greek
set's five of ten. English success falls from $96.4/95.6/96.6\%$ to $93.8/93.4/87.2\%$,
drops of $2.6$, $2.2$ and $9.4$ points, a mean of $4.7$ against $23.3$ in Greek. Every Greek
drop still exceeds every matched-English drop, though the margin is now $14.4$ against $9.4$
rather than the comfortable gap the syntax-only control suggested.

Two things follow, and we state both. Substituting the referring noun is \emph{not} free in
English: one seed lost $9.4$ points, so part of what the Greek set costs is the size of the
perturbation and not the language. But English absorbs the harder perturbation at $4.7$ points
where Greek loses $23.3$ under a milder one, roughly five times more, and the English cost
barely moves as the perturbation goes from touching no nouns ($3.9$) to touching seven of ten
($4.7$) while Greek at an intermediate strength is five times worse than either. Generic
brittleness to rewording therefore cannot explain the Greek drop. What we claim is that
bounded statement, not that language accounts for all of it.

Two conclusions follow, and the first is uncomfortable. \emph{A large fraction of
the headline number is translator-specific}: the three seeds lose $63$, $31$ and
$26\%$ of their trained-phrasing level when the translator changes, a mean of
$40\%$, against $5\%$ for a rewording in English that substitutes the referring noun more
often than the Greek set does. A policy trained on machine-translated
instructions learns, to a substantial degree, the phrasing habits of the system
that produced them; reported in the target language's name, this overstates what
was learned. Anyone localizing a stack this way should evaluate with an independently produced
instruction set. We previously reported that paraphrases from our own generator understate
this gap by more than ten points, and we withdraw that: it held at the seed we first
measured and does not survive replication. Across three seeds our own held-out paraphrase
sets cost $23.8$ and $23.5$ points against the independent set's $23.3$, which is the same
penalty, not a smaller one. What the independent set buys is therefore not a larger measured
drop but an instrument we did not author; the argument for using one is provenance, not
magnitude, and the penalty is not a model copying one translator's words but brittleness to
any unseen Greek phrasing. Second, what remains is real: every one of the six policies we
measured across both recipes scores above its instruction-blind floor on the
independent set, so the policies do follow Greek they have never seen phrased that
way. The honest headline is therefore not ``48.6\% Greek instruction-following''
but ``Greek instruction-following that loses roughly a third of its value when the
translator changes.''

\begin{figure*}[t]
\centering
\includegraphics[width=0.92\textwidth]{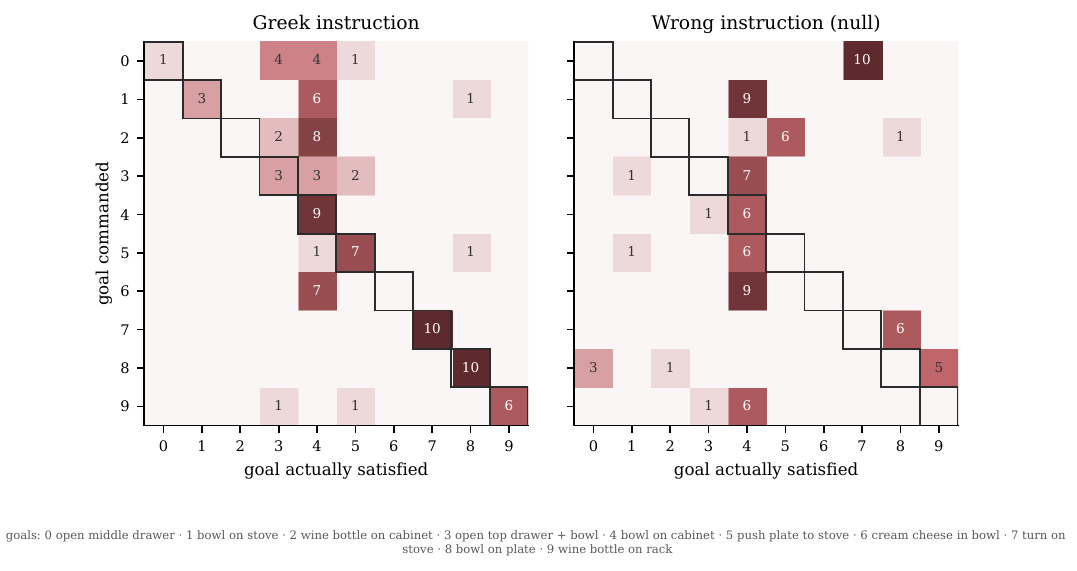}
\caption{What the policy does when it fails. Rows are the commanded goal, columns
the goal actually satisfied at episode termination; ringed cells are the diagonal,
where the policy did what it was told. Under Greek instructions the diagonal is
populated but so is one column: goal~4 absorbs most of the off-diagonal mass. Under
the wrong-instruction null the diagonal nearly empties while that same column
persists, so the fallback is one over-practised behaviour rather than random
action. Greek failures are therefore comprehension failures with a specific
attractor, not motor failures. One policy, one seed, 100 episodes per condition.}
\label{fig:confusion}
\end{figure*}

\paragraph{What failure looks like.} Success rates say how often a policy
succeeds, never what it does instead. We therefore scored all ten goal predicates
at the end of every episode, for a hundred episodes per condition, which
distinguishes a policy that attempted the right task and fumbled it from one that
competently performed the wrong one. The two languages fail in different ways.
Of the English failures, \emph{none} ended with any goal satisfied: the policy
tried the commanded task and missed. Of the Greek failures, $82\%$ ended with the
robot having successfully completed a \emph{different} task in the scene. The
wrong-instruction control fails the same way, at $79\%$ (Figure~\ref{fig:confusion}). Greek failure is
therefore not a motor deficit but a comprehension one, and its signature is
statistically indistinguishable from having been given an unrelated command.

The misdirection also has a single attractor. One goal, \emph{put the bowl on top
of the cabinet}, absorbs $29$ of the $42$ Greek diversions and $38$ of the $74$
wrong-instruction ones, so the fallback is one over-practised behaviour rather
than random flailing; it is also among the goals the policy performs best when
commanded. The clearest case is again a minimal pair: commanded \emph{put the wine
bottle on top of the cabinet} in Greek, the policy performs the bowl variant eight
times, preserving the destination and substituting the object. This is the
referent-collapse signature above, now read from a mechanism rather than inferred
from an aggregate. We report it for one policy at one seed, so the attractor's
identity should not be assumed stable across runs; the asymmetry between motor and
comprehension failure is the transferable part.

\begin{figure}[t]
\centering
\includegraphics[width=\columnwidth]{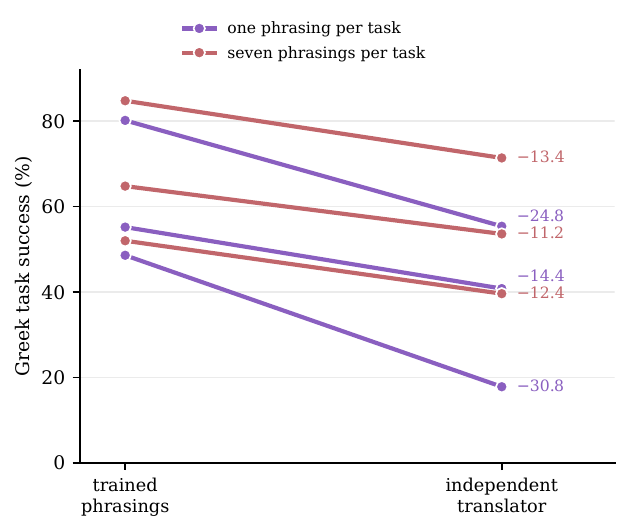}
\caption{The translator penalty, measured within each policy. Every line is one
training run moving from the phrasings it was trained on to an independently
produced translation of the same commands. All six fall, so the penalty is real;
the slopes separate perfectly, so phrasing diversity halves it. Note that the
\emph{levels} overlap between the two arms while the \emph{drops} do not: seed
variance is larger than the effect, and only the paired comparison cancels it.}
\label{fig:idiolect}
\end{figure}

\paragraph{Phrasing diversity halves the translator penalty.} Because the
preceding policies see a single Greek phrasing per task, we retrained with seven
distinct Greek phrasings per task while holding the language sampling ratio
fixed, isolating phrasing variety from the amount of target-language gradient.
Comparing the two recipes by their success \emph{levels} does not work here: with
three seeds each they overlap ($39.6$--$71.4\%$ against $17.8$--$55.4\%$ on the
independent set), because seed variance is larger than the effect.
Comparing each policy against itself does work, because both terms of a
within-policy drop come from one checkpoint and the seed term cancels (Figure~\ref{fig:idiolect}). Measured
that way the effect is unambiguous: moving to the independent translator costs the
single-phrasing policies $30.8$, $24.8$ and $14.4$ points, and the diverse policies
$12.4$, $11.2$ and $13.4$. Every single-phrasing drop exceeds every diverse one, an
exact run-level permutation test returns $p = 0.100$ (the smallest value three runs
per arm can produce), and the spread of the penalty collapses from $16.4$ points to
$2.2$. Diversity roughly halves the penalty and makes it predictable. It is the only
intervention we tried that survives replication, and it is worth the cost we also
measure: English success falls $3.6$ points, 95\% CI $[-6.4, -0.8]$. This is the cost of
\emph{training} on seven phrasings, and should not be confused with the $3.9$-point cost of
\emph{evaluating} on an unseen English phrasing reported below.

\begin{figure}[h]
\centering
\includegraphics[width=\columnwidth]{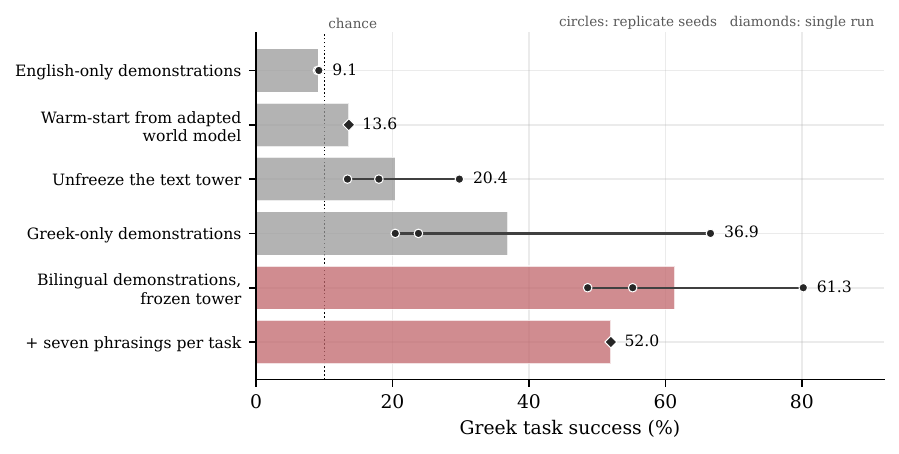}
\caption{Greek task success by training intervention. Points are individual
training runs; the arms we replicated carry three seeds each. Only two
configurations separate from the bilingual baseline once seed variance is
included: removing target-language demonstrations entirely (which lands on its
own wrong-instruction floor) and unfreezing the text tower. The Greek-only arm
overlaps the baseline and is reported as unresolved. Task-level intervals, shown
for single-run arms, understate uncertainty for comparisons \emph{between}
policies, because two runs of one recipe differ by seed alone.}
\label{fig:interventions}
\end{figure}

\begin{figure*}[t]
\centering
\includegraphics[width=\textwidth]{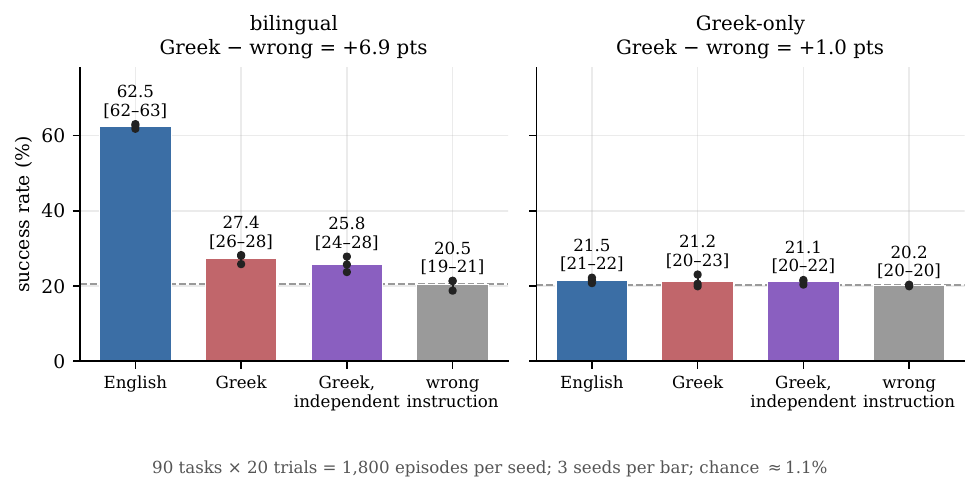}
\caption{The same four-condition protocol on a suite nine times larger, three training
seeds per arm. Bars are seed means, dots are seeds, whiskers the seed range. Ninety tasks
drop uniform chance from $10\%$ to $1.1\%$ and remove the ten-cluster ceiling that bounds
every interval elsewhere in this paper. Uniform chance is not the relevant baseline: the
suite is twenty scenes of two to seven goals, so scene recognition alone scores about
$22\%$, and the dashed line is the measured wrong-instruction floor we read margins
against. The bilingual policy keeps a graded response on every
seed: correct Greek above an independent Greek rendering, both above the wrong-instruction
control, by a margin that varies only $0.3$ points across seeds ($6.7$ to $7.1$). The
Greek-only policy's margin over its own control never exceeds $2.7$ points and averages
$1.0$: given correct Greek, Greek phrased by a translator it never saw, a wrong
instruction, or English it was never trained on, it does close to the same thing.}
\label{fig:scale90}
\end{figure*}

\paragraph{Locating the residual gap.} Three probes test the obvious explanations
for why Greek trails English (Table~\ref{tab:policy}, lower half; Figure~\ref{fig:interventions}), and all three
fail in the same direction. Training on English alone yields 98.2\% English and
9.0\% Greek: chance. A multilingual tower transfers \emph{nothing} to the action
pathway without demonstrations in the target language, and bilingual training
costs only 1.8 points of English for 39.6 points of Greek. Training on Greek
alone, which doubles Greek gradient share from one half to all of it, scored
20.4\% in our first run, and we initially read that $28$-point deficit as
evidence that English demonstrations scaffold Greek. Replication withdraws it.
Across three seeds the Greek-only arm scores 20.4/66.6/23.8\% against the
bilingual arm's 48.6/80.2/55.2\%: the ranges overlap, the best Greek-only run
beats two of three bilingual runs, and an exact run-level permutation test gives
$p = 0.300$ where $0.100$ is the smallest value three runs per arm can attain.
The means still favour the mixture (61.3\% vs.\ 36.9\%) and Greek-only is the
most seed-unstable arm we measured (46.2-point range), so we report a direction
this design cannot resolve rather than a mechanism. What does replicate is that
every Greek-only run reads its instructions on this suite (20.4/66.6/23.8\% against 2.2/3.6/8.4\%
under wrong ones). On the ninety-task suite the same recipe clears its floor by at most $2.7$ points on any
of three seeds (Figure~\ref{fig:scale90}); we return to the point below.
Finally, unfreezing the text tower, so that it may adapt to Greek directly,
degrades both languages (82.4\% / 18.0\%) despite normal convergence; small-data
action training damages the pretrained multilingual representation rather than
improving it.

These failures were each significant under a task-level test, and we resist
unifying them; replication later showed that test to be the wrong one for
comparing policies, and withdrew one of the three (Section~\ref{sec:study2}). An earlier draft of this paper proposed the slogan
``do not touch the tower''; auditing our own configurations refuted it. The
world model whose checkpoint we warm-start from was itself trained with the
tower frozen (only the generation pathway and its projections were in the
optimizer), and the monolingual probes likewise use the stock frozen-tower
recipe. In two of the three failing interventions the multilingual
representation is bit-identical to the one in the winning recipe. The failures
therefore have three distinct locations: warm-starting damages the
\emph{generation pathway} that the policy recipe also trains; monolingual
training changes the language composition of the \emph{data}; and only the
third actually perturbs the tower. Of the three, only the tower intervention and
the no-Greek ablation survive seed replication as separable from the baseline. We report them separately, and note that the
tower result replicates across three seeds but tests a single \emph{configuration},
maximal aggression (every attention and
MLP projection, the embedding table, and both layer norms unfrozen at a flat
learning rate), which says nothing about gentler adaptation such as low-rank
adapters or a strongly damped tower learning rate.

We can therefore say what did not work and where, but our evidence does not
identify the cause of the residual English--Greek gap. Three probes, two of them
replicated and one at a single run, do not close an intervention space. Two explanations remain open and we cannot
separate them. The first is representational: the tower encodes Greek less
richly than English, and fine-tuning at this scale cannot supply what pretraining
did not. The second is mechanical, and cheaper to fix: the tower's tokenizer
fragments these Greek instructions into $3.81\times$ as many tokens as their
English originals, most of them single characters, so the action pathway must
learn goal selection from twenty-four sub-lexical pieces where English offers
eight words. That would also predict the failure we observe, since
character-level segmentations of \emph{the bowl on top of the cabinet} and
\emph{the wine bottle on top of the cabinet} share most of their token
$n$-grams. Distinguishing the two inside one tower requires either a direct probe of the frozen
tower's encodings or a retrained policy on transliterated or
vocabulary-extended Greek; both are future work.

Comparing \emph{across} the two towers, however, is free, and it cuts against the
mechanical explanation. The English-centric 2B tower, the one that never learns Greek at
all in Study~1, fragments the same ten Greek instructions \emph{less} than the
multilingual 8B tower that learns it: $150$ tokens against $274$, an inflation over
English of $2.08\times$ rather than $3.81\times$, and $0.45$ tokens per character rather
than $0.81$. If fragmentation were what blocks a language, the tower with the coarser
Greek segmentation should be the more teachable one, and it is the opposite. We take this
as evidence that fragmentation is not sufficient to explain a tower's failure to acquire
a language. It does not establish the converse, that representational depth carries the
weight instead: the two towers differ in parameter count and in pretraining corpus as well
as in fertility, so this is one counterexample against a sufficiency claim, not an
attribution of cause at $n=2$. It does not settle the residual gap \emph{within} the multilingual tower,
where the two remain entangled. We report the diagnostic because it costs nothing: anyone
choosing a base model can compute both towers' fertility in the target language before
spending a GPU-hour, and should not read a low count as a green light.

\paragraph{Does any of this survive a larger suite?} Every number above rests on ten
goals, and ten goals is ten clusters. We therefore repeated the protocol on the
ninety-task suite, both arms, three training seeds each, four instruction conditions,
$20$ trials per task, so $1{,}800$ episodes per cell and $43{,}200$ in total. Uniform
chance falls from $10\%$ to $1.1\%$, but that figure misdescribes the problem the policy
faces: the ninety tasks are twenty scenes carrying two to seven goals each (mean $4.5$),
so a policy that recognises the scene and performs a goal afforded by it scores about
$22\%$ without reading the instruction. The measured wrong-instruction floor is $20.5\%$,
$1.7$ points below that, which is why we read margins over the measured floor throughout
rather than distance from uniform chance.
Three things change, and only the first is comfortable.

\emph{The bilingual policy reads Greek on every seed, and the margin is what shrinks.} It
scores $27.4\%$ under correct Greek ($25.8/28.2/28.1$ across seeds) against a $20.5\%$
wrong-instruction floor ($18.8/21.4/21.3$), a separation of $6.9$ points that varies by
only $0.3$ points across seeds ($7.1/6.8/6.7$; Figure~\ref{fig:scale90}). The effect is
real and stable and the ordering is intact, but Greek success as a fraction of English
falls from roughly one half on ten goals to $44\%$ here ($27.4\%$ against $62.5\%$), and
the absolute English--Greek gap widens to $35$ points. Both ratios count the
wrong-instruction floor as Greek following. Corrected for it, Greek retains
$16\%$ of English instruction-following at ninety tasks
($(27.4-20.5)/(62.5-20.5)$), and we prefer that number. A reader who took ``roughly half of English performance'' from our
ten-goal result should read it as an upper bound obtained on an easier suite.

\begin{figure*}[t]
\centering
\includegraphics[width=\textwidth]{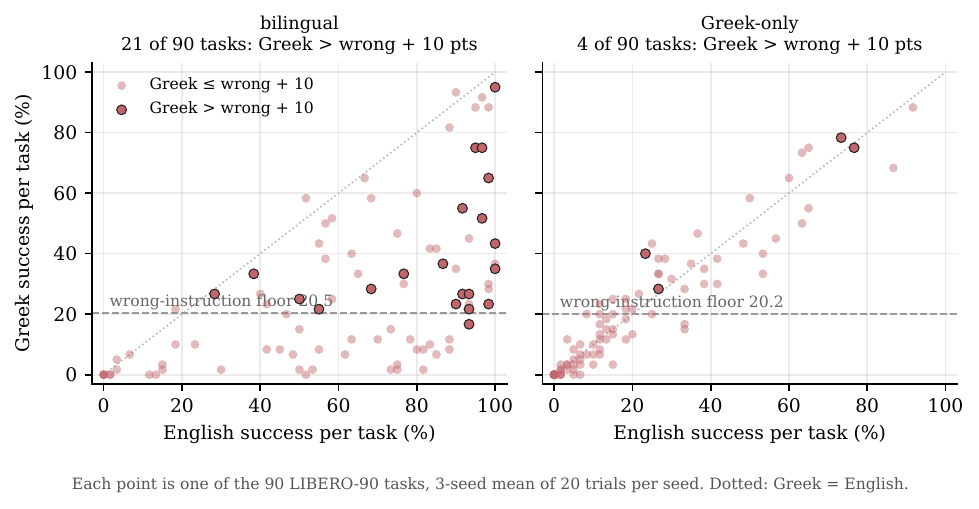}
\caption{Where Greek works, task by task. Each point is one of the $90$ tasks, three-seed
mean of $20$ trials per seed; the dashed line is that arm's wrong-instruction floor and the
dotted diagonal is Greek equal to English. Outlined points are tasks on which Greek clears
the task's own wrong-instruction rate by more than ten points. The bilingual policy reads
Greek on a minority of tasks and rides the motion prior on the rest. The Greek-only policy's
Greek and English success move together almost perfectly (correlation $0.95$), because
neither instruction is changing what it does.}
\label{fig:pertask90}
\end{figure*}

\emph{The Greek-only policy barely reads at all, on any seed.} Its margin over its own
wrong-instruction floor is $-0.3$, $+0.7$ and $+2.7$ points across the three seeds (mean
$1.0$), and its four conditions span $0.8$, $1.7$ and $2.7$ points. On ten goals this arm
cleared its floor on all three seeds, by $18$ to $58$ points, and we read that as weak but
genuine instruction-following; here the largest margin any seed produces is $2.7$. Every
bilingual margin (minimum $6.7$) exceeds every Greek-only margin (maximum $2.7$): perfect
separation at three seeds per arm, for which an exact run-level permutation test returns
$p=0.100$, the floor of this design rather than a small $p$. The stronger evidence is
within each run, where ninety tasks give a real interval. A paired task-level bootstrap
over the ninety tasks puts the bilingual margin at $+7.1$ $[+3.8, +10.4]$, $+6.8$
$[+3.4, +10.4]$ and $+6.7$ $[+3.6, +10.1]$ across seeds, excluding zero in all three. The
target-only arm gives $-0.3$ $[-1.7, +1.1]$, $+0.7$ $[-1.2, +2.4]$ and $+2.7$
$[+1.3, +4.3]$: zero falls inside the interval for two of its three seeds and outside for
the third, so the honest statement is that target-only training reads Greek weakly and
inconsistently, not that it never reads it. Because both terms of each margin come from
one checkpoint, the seed term cancels within a policy, so this compares within-policy
quantities rather than levels. It is the strongest evidence we have that
target-language demonstrations alone are not sufficient. It is weaker evidence for the
stronger reading, that the English demonstrations are what make Greek legible: that is a
between-policy claim, and a within-policy statistic constrains it only through the
ordering of the two arms' margins. In this paired form
and on this suite it reinstates the scaffolding reading we withdrew on ten goals; we still
do not report it as a level. The per-task view says the same thing in a different way
(Figure~\ref{fig:pertask90}): for the bilingual policy, Greek success clears the task's own
wrong-instruction rate by more than ten points on $21$ of $90$ tasks and correlates with
English success at $0.57$; for the Greek-only policy it does so on $4$, and its Greek and
English per-task success correlate at $0.95$, which is what a policy that does not read the
instruction language looks like.

\emph{The translator-idiolect penalty largely disappears, and we do not think that is
good news.} Moving the bilingual policy from the training translator's phrasings to an
independently produced Greek rendering costs $2.1$, $0.4$ and $2.3$ points across seeds
(mean $1.6$) here, against $23.3$ points on ten goals. The tempting reading is that idiolect matters less at scale. The likelier one
is that the instrument has run out of room: on ten goals Greek sat $44$ points above its
floor, and a penalty had somewhere to go, whereas here it sits $6.9$ points above and
cannot fall far before hitting the motion prior. A penalty measured against a compressed
dynamic range is not evidence of a smaller effect. We report the number and decline to
interpret it as a reduction.

Two further things follow. First, the seed instability that dominates the ten-goal
results is largely a small-suite artifact: Greek success for the bilingual recipe varies
by $2.4$ points across seeds on ninety tasks where it varied by $31.6$ on ten. Ninety tasks
buy not only a tighter estimate of each policy but a stable one. Second, the suite is no
longer what binds this paper; what binds every cross-policy claim is now the floor of the
run-level test at three seeds per arm ($p=0.100$), and pushing below it costs one training
run per seed per arm.

\section{The Real-Robot Corpus We Cannot Score}
\label{sec:droid}

Ninety-one percent of our translated instructions belong to
DROID~\cite{khazatsky2024droid}: $53{,}096$ unique instructions carried by
$57{,}639$ success episodes of real teleoperation,
against LIBERO's $111$ tasks. It is by far the larger asset, and every number in
this paper comes from the smaller one. The reason is not that the data is
unusable but that we cannot \emph{score} it. DROID is real-robot data with no
simulator, so the closed-loop success protocol on which every result above rests
requires a physical Franka arm we do not have.

We report the position honestly rather than omitting the corpus. A bilingual
Greek policy \emph{is} trained on it, using the recipe this paper's evaluations
selected: from base, tower frozen, an equal mixture of English and Greek
instructions. Instruction-level holdout is enforced rather than episode-level,
because the same command recurs across episodes and an episode split would leak
held-out phrasings back into training through a different episode. The accounting is:
$57{,}639$ success episodes carry $53{,}096$ unique instructions; $1{,}000$ of those
instructions, appearing in $1{,}423$ episodes, are held out; $56{,}216$ episodes remain for
training. The $53{,}207$ figure quoted in Section~\ref{sec:setup} is the translated
instruction set, $53{,}096$ from this corpus plus $111$ from LIBERO, and is a count of
strings rather than episodes. Absent a simulator, the available proxy is
action-prediction error on held-out episodes under the same three-way contrast
used throughout, and we require it to clear its own gate before reading anything
else from it: if it cannot separate a correct English instruction from an
unrelated one, it is not measuring language and none of its Greek values mean
anything.

It clears the gate, and the result is negative for Greek. Over $150$ held-out
windows per condition, English instructions predict the ground-truth action chunk
better than an unrelated instruction by $2.28\%$ (MSE $0.290$ against $0.297$), from a
single run at $150$ windows per condition and reported without an interval; we read the
gate as directional evidence that the metric responds to language at all, not as a
measured effect size.
Against that working metric, Greek is indistinguishable from the null: $0.297$,
a separation of $0.03\%$. An independently produced Greek translation scores
$0.65\%$ \emph{worse} than the null. The gate passing is what makes this
readable. Had every condition been flat, a blind metric and an untrained policy
would be indistinguishable; because English separates on the same checkpoint, the
instrument is working and the flat Greek is a property of the policy.

We state the scope carefully. The English signal is itself small, and this
checkpoint is $500$ iterations of a recipe written for $10{,}000$, so instruction
conditioning is only beginning to emerge in either language. The claim we can
defend is that \emph{at five percent of the intended schedule, English
conditioning is already detectable on real-robot data and Greek conditioning is
not}. Whether Greek emerges later is untested and would cost roughly three days of
compute to answer on a proxy that still cannot produce a success rate; we judged
that a poor trade and say so rather than leaving the reader to assume we ran it.

Two things follow for anyone repeating this. First, the translation is cheap and
the evaluation is not: at a cost of hours we produced Greek for $57{,}639$
real-robot episodes, ten times every simulated suite we had combined and a hundred
and thirty times the one our results come from, and then measured none of it. The
bottleneck was never the language. Second, a proxy metric needs its null before it needs its result, which
is the same discipline the wrong-instruction control enforces above and the same
one a colour-histogram metric evaded twice
(Section~\ref{sec:lessons}). We describe the held-out construction and the
independent re-translation in enough detail to be repeated, so the measurement can
be done by anyone with the hardware and a translation pipeline of their own.

\section{Lessons for Evaluation}
\label{sec:lessons}

\begin{figure*}[t]
\centering
\includegraphics[width=0.86\textwidth]{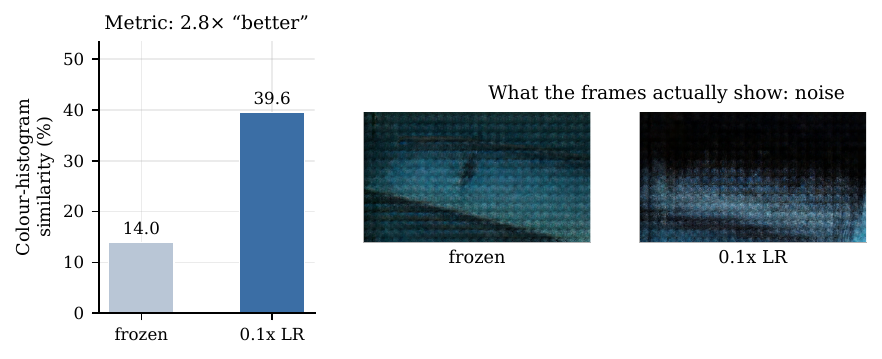}
\caption{An instrument failure. A colour-histogram similarity metric reported a
2.8-fold improvement in Greek-conditioned generation between two fine-tuning
configurations; both outputs are structureless noise whose palette had drifted
toward the reference video.}
\label{fig:metric}
\end{figure*}

Five instruments failed us informatively. First, a color-histogram similarity
metric against reference videos doubled, twice, while the underlying generations
remained pure noise whose palette had drifted toward the reference (Figure~\ref{fig:metric}); coherence
must be judged structurally or by a calibrated judge, never by color statistics.
Second, single-goal benchmarks silently cannot falsify language claims; the
wrong-instruction control costs one extra evaluation run and would have exposed
the 84.6\% Greek illusion immediately.

Third, and the instrument that cost us the most, a single training run per arm is
not a measurement in the low-resource language. Retraining one recipe with only
the seed changed moves Greek success across $48.6/80.2/55.2\%$ while English moves
a single point. Any cross-recipe difference smaller than about thirty points is
therefore unresolvable at one run per arm, and three of our five retractions are
of exactly that shape: an effect that looked clean at $n=1$ and vanished when the
arm was replicated. The asymmetry is the practical warning. Nothing in our English
numbers would have suggested this instability existed, so a practitioner
validating a pipeline in the high-resource language will conclude, wrongly, that
single runs are informative.

\begin{figure}[t]
\centering
\includegraphics[width=\columnwidth]{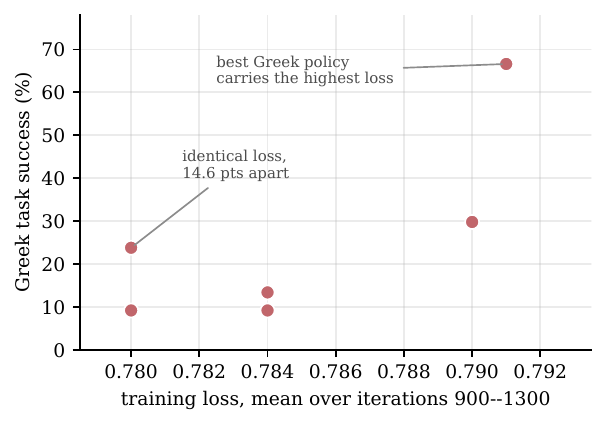}
\caption{Training loss cannot rank policies on the property this paper measures.
Six completed policies: loss spans $1.4\%$ of its mean while Greek success spans a
factor of $7.2$. Behaviour-cloning loss is dominated by action prediction from the
visual prior, so language conditioning is too small a term to surface in it.}
\label{fig:lossblind}
\end{figure}

Fourth, and most cheaply overlooked: \emph{training loss carries no usable signal
about instruction-following} (Figure~\ref{fig:lossblind}). Across six completed policies the late-training loss
spans $0.780$--$0.791$, a range of $1.4\%$, while their Greek success spans
$9.2$--$66.6\%$, a factor of $7.2$. Two policies at identical loss ($0.780$) differ
by $14.6$ points of Greek, and the best Greek policy of the six carries the
\emph{highest} loss. The cause is structural rather than incidental:
behaviour-cloning loss is dominated by action prediction from the visual and
proprioceptive prior, so language conditioning is a small enough term that a policy
at chance in the target language and one at $66.6\%$ are indistinguishable to the
optimiser. This is the same blindness as the single-goal suites, one layer up, and
it means a training curve can never be used to rank recipes on the property this
paper is about.

\begin{figure}[t]
\centering
\includegraphics[width=0.9\columnwidth]{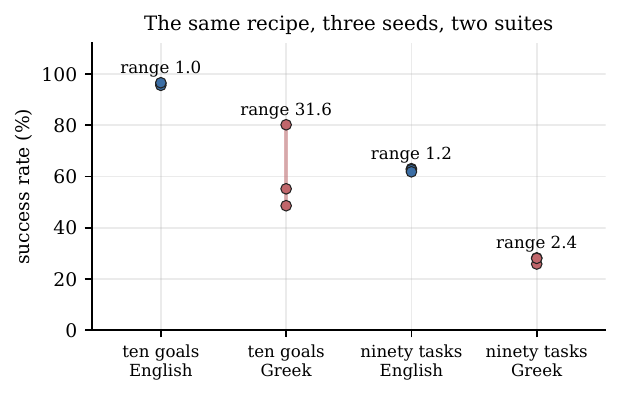}
\caption{The seed instability is a property of the small suite, not of the recipe. The same
bilingual recipe at three seeds: Greek success spans $31.6$ points on ten goals and $2.4$ on
ninety tasks, while English spans about one point on both. A single-run Greek number on a
ten-goal suite is a draw from a thirty-point range; on ninety tasks it is close to a
measurement.}
\label{fig:seedstab}
\end{figure}

Fifth, a small multi-goal suite is itself an instrument, and it misled us in two ways
we saw only by scaling it. On ten goals the Greek-only policy cleared its
wrong-instruction floor on every seed and we read that as weak but genuine
instruction-following; on ninety tasks, three seeds per arm, its margin is at most
$2.7$ points. And the $31.6$-point seed instability we had treated as a property of
low-resource training is largely a property of ten clusters: the same recipe varies by
$2.4$ points across seeds on ninety tasks (Figure~\ref{fig:seedstab}). Ten goals does not
only bound every interval; it manufactures variance that a larger suite removes.
The same caveat we apply to the shrunken idiolect penalty applies here in the
authors' favour and we state it too: a quantity that sits $6.9$ points above its
floor has less room to vary than one that sits $44$ above, and the two suites were
trained on different data, so part of this contraction is range, not stability.

What the five share is that each reports success where there is none, and each is
cheaper to consult than the control that exposes it. That is why we pair every
positive claim in this paper with a condition under which it would have failed:
success-rate parity is not understanding, and five of our own preliminary
conclusions were retracted when a control or a replication contradicted them;
Table~\ref{tab:retractions} lists them.

\begin{table*}[t]
\centering
\footnotesize
\setlength{\tabcolsep}{6pt}
\renewcommand{\arraystretch}{1.30}
\begin{tabular}{@{}L{0.30\textwidth}L{0.32\textwidth}L{0.32\textwidth}@{}}
\toprule
\textbf{Claim we drew} & \textbf{What refuted it} & \textbf{Status now} \\
\midrule
``Do not touch the tower'': every intervention that helped left it frozen &
Audit of our own configurations: two of the three failing interventions leave the tower
bit-identical &
Withdrawn as a unifying rule; the tower result survives on its own \\
English demonstrations scaffold Greek (bilingual $48.6$ vs target-only $20.4$) &
Seed replication: ranges overlap, run-level $p=0.300$ &
Withdrawn as a level; reinstated as a paired margin comparison at ninety tasks
(Section~\ref{sec:study2}) \\
Phrasing diversity raises the independent-translator \emph{level} ($39.6$ vs $17.8$) &
Seed replication: $39.6$--$71.4$ against $17.8$--$55.4$, overlapping &
Withdrawn as a level; survives as a within-policy drop \\
Held-out-paraphrase success decays monotonically with distance from the training
translator &
Seed replication: the ordering holds in two of five paired comparisons and inverts by
$36$ points in one &
Withdrawn; we report that phrasing change costs accuracy, not how much \\
Our own paraphrases understate the translator gap by more than ten points &
Seed replication: own-generator drops average $23.8$ and $23.5$ against the independent
set's $23.3$ &
Withdrawn; the case for an independent instrument is provenance, not effect size \\
\bottomrule
\end{tabular}
\caption{The five retractions. Each was drawn from a single run or a single seed, and each
was refuted by a control or a replication rather than by an outside reader. The second
returns in a weaker, paired form; we count it as retracted because the level it originally
asserted does not stand.}
\label{tab:retractions}
\end{table*}

\section{Limitations}
\label{sec:limits}

One language, one model family, simulation only. The Greek corpus is machine generated,
rephrased rather than literally translated, and carries machine-artifact risk. \emph{No human-authored Greek appears anywhere
in this study.} Every Greek string a policy trained on or was evaluated against was written
by a language model, including the ``independent'' set: independence there means a different
model with a different, glossary-free prompt, not a native speaker. This bounds the
idiolect result specifically. What we measure is the cost of moving between two machine
renderings of the same commands, which is a lower bound on the cost of moving to Greek as a
person would actually speak it, and it leaves open the possibility that both renderings share
machine-translation artifacts that a human set would not. Collecting native phrasings for
these ten commands is an afternoon's work for a Greek speaker and is the first thing we
would add. World-model binding was judged by a
single VLM judge (calibrated, but one model). Policies generalize partially to unseen phrasings of
their trained goals, well above the wrong-instruction floor in every run, but the
level is strongly seed-dependent (24--62\% across three seeds on one paraphrase
set) and we can only report it as a range. Every goal a policy is asked to perform
was seen in training; we measure no generalization to novel \emph{tasks}. We do not identify the cause of the residual
Greek--English gap: our probes eliminate three explanations but leave
representational depth and tokenizer fragmentation ($3.81\times$, measured)
entangled, and we cannot say which dominates. Ten goals is ten clusters, and it binds every interval we
report on that suite; we therefore repeated the protocol on ninety tasks, which removes
that ceiling and changes two of our readings (Section~\ref{sec:study2}). What binds the
paper now is the run-level test rather than the suite: the ninety-task arms are three
training runs each, enough for perfect separation between recipes but not to push an exact
permutation test below its $0.100$ floor. Each further seed costs one training run per
arm, and that is the most valuable next step for anyone repeating this. Our cross-policy comparisons originally rested on one
training run per arm against a measured Greek seed spread of over thirty points.
We have since replicated four arms at three seeds each, which resolved two
comparisons and dissolved one; the arms still at $n=1$ (the warm-start policy, and
the world-model dose ladder of Study~1) remain directional, and we mark them as
such wherever they appear.

\paragraph{What remains open.} We state these as questions rather than
retiring them, because each is answerable and none is answered here.
\emph{(i)~Does the language mixture matter beyond the presence of target-language
data?} Bilingual training beats target-only on average across three seeds each
($61.3\%$ against $36.9\%$) but the ranges overlap and an exact run-level test
returns $p=0.300$, where $0.100$ is the floor at this design; we first read the
single-run version of this comparison as evidence that the high-resource language
scaffolds the low-resource one, and withdrew that reading. The ninety-task suite at three
seeds per arm answers it in paired form: every bilingual seed's margin over its own
wrong-instruction floor ($6.7$ to $7.1$ points) exceeds every target-only seed's ($-0.3$
to $2.7$), $p=0.100$ at this design, and the Greek levels no longer overlap ($25.8$ to
$28.2$ against $19.9$ to $23.1$). On this evidence the mixture matters, though at a design
whose run-level test cannot go below $p=0.100$ we state it as a direction rather than a
demonstrated effect. What stays open is how much: three
seeds per arm cannot push the run-level test below $0.100$, and we do not report a level
difference. \emph{(ii)~Does any of this transfer beyond the
simulator?} Our translated corpus is dominated by a real-robot dataset that no
experiment here consumes, because closed-loop success on it requires physical
hardware; the action-prediction proxy on held-out episodes is reported in
Section~\ref{sec:droid} and is negative for Greek at five percent of the intended
training schedule. \emph{(iii)~What sets the residual target--English gap?} Our
probes eliminate three explanations but leave representational depth and
tokenizer fragmentation entangled. \emph{(iv)~Does the phrasing-diversity benefit
extend past ten goals?} It replicates cleanly as a within-policy effect on the ten-goal
suite. Our ninety-task runs do not test it: neither arm is phrasing-diverse, and building
one needs a diverse ninety-task training set we have not built. It remains untested, and
the compressed idiolect penalty we measure at ninety tasks makes it harder, not easier,
to detect.

\section{Conclusion}

\begin{table*}[t]
\centering
\small
\setlength{\tabcolsep}{5pt}
\renewcommand{\arraystretch}{1.20}
\begin{tabular}{@{}r@{\hspace{7pt}}L{0.35\textwidth}L{0.46\textwidth}c@{}}
\toprule
 & \textbf{Do} & \textbf{What it bought, measured} & \textbf{Runs} \\
\midrule
1 & \textbf{Check that the base model's text tower already handles the language.} It is free to test and decides the project. Tokenizer fertility is a cheap first look but not a sufficient test. & English tower: noise from Greek on every rung of a dose ladder. Multilingual tower, identical recipe: coherent Greek scenes on 64\% of prompts. The failing tower fragments Greek \emph{less} ($2.08\times$ against $3.81\times$). & 1 each \\
\addlinespace[3pt]
2 & \textbf{Train the policy from the base checkpoint, not from a language-adapted world model.} & Warm start 79.2 / 13.6\% (EN / EL); from base 96.4 / 48.6\%. & 1 \\
\addlinespace[3pt]
3 & \textbf{Append the translated instructions as pipe-separated variants beside the English ones.} The loader samples 50/50; no code changes. & Greek margin over the wrong-instruction floor 6.7--7.1 points on every seed; Greek demonstrations alone: at most 2.7. & 3 per arm \\
\addlinespace[3pt]
4 & \textbf{Leave the text tower frozen.} & Unfrozen: 82.4 / 18.0\%, worse in both languages, replicated. & 3 \\
\addlinespace[3pt]
5 & \textbf{Train on several phrasings per task, not one.} & Independent-translator penalty 23.3 $\to$ 12.3 points; its spread 16.4 $\to$ 2.2. & 3 per arm \\
\addlinespace[3pt]
6 & \textbf{Evaluate on a suite where each scene admits several goals, and always run the wrong-instruction control.} & Single-goal suite: 84.6\% Greek against 82.6\% wrong. The control costs one evaluation run. & n/a \\
\addlinespace[3pt]
7 & \textbf{Evaluate on an independently produced translation.} Run the same rewording in the high-resource language as a control, matched for how much it changes. & Our own paraphrases cost the same as the independent set ($23.8$, $23.5$ vs $23.3$), so the case for an independent set is provenance, not a larger number. English rewording costs $3.9$ points with syntax alone and $4.7$ when it also substitutes the referring noun in seven of ten commands, against $23.3$ in Greek. & 3 \\
\addlinespace[3pt]
8 & \textbf{Train at least three seeds per arm; compare within-policy margins, never levels.} & Greek moved 31.6 points on the seed at ten goals (2.4 at ninety); English moved 1.0. & n/a \\
\bottomrule
\end{tabular}
\caption{The recipe, in the order we would apply it, with the measurement behind each
step and the number of training runs that support it. Steps resting on one run are
directional; the others replicated across seeds.}
\label{tab:recipe}
\end{table*}

Adding a language to a robot foundation model is cheap where the base model's
tower already speaks it, and out of reach of every caption fine-tuning
configuration we could afford where it does not.
Given the right substrate, a corpus of machine-produced Greek and a data change that
requires no code carry a policy from chance to roughly half of English performance on a
ten-goal suite and two fifths of it on a ninety-task one, but only with demonstrations in
\emph{both} languages: target-language data is necessary, and on the larger suite a
policy trained on it alone barely follows it, under three points over its own control
on every seed. The two
interventions that look most promising, initializing from a language-adapted
world model and adapting the tower directly, both hurt.

What survives of the target language is partial and non-compositional, though less
uniformly than we first reported. Greek success is sharply bimodal across goals
\emph{at the seed we profiled}: five goals at $0.72$--$0.98$ and five at $0.00$--$0.24$.
Profiling all three seeds shows the split is not a property of the recipe
(Appendix~\ref{app:pergoal}). The per-goal counts above $50\%$ are $5$, $9$ and $6$ of ten,
and only one goal (goal~0) stays below it in every seed, so the identity of the failing
goals otherwise moves with the seed while four goals succeed in every one. We could not identify a lexical property separating the modes:
every instruction in the set shares a content noun with at least one other. We would not have seen
that structure from aggregate success rates, and we would not have trusted the
aggregates at all without a control that current single-goal benchmarks make
mandatory. For anyone localizing such a stack, the practical order
(Table~\ref{tab:recipe}) is: check the tower first, since it is free to test and decides the project; supply
demonstrations in the target language, since nothing else substitutes for them, and keep
the high-resource ones alongside, since on our larger suite the target language alone
bought almost nothing;
leave the pretrained representation alone; and measure with a null you can
guarantee rather than one you assume.

We close on the part we expect to outlast the numbers. Language adaptation of
vision-language-action models was, when we began, thinly covered from where we could
see: the multilingual literature largely stops where the output is text, and the robotics
literature is written in English throughout. We expect that to change quickly. That means the first groups to work
here will be calibrating instruments rather than beating baselines, and will do it
without a body of prior negative results to warn them. This study is an attempt to
supply some. Of the conclusions we drew over its course, five did not survive a
control or a replication, and each failed in a way that would have been invisible
to the measurement a reasonable practitioner would have chosen first. We would rather publish that record than a
cleaner one we trust less, and we think the controls that produced it transfer
further than any of our numbers do.

\section{Availability}
The policy studied in Section~\ref{sec:study2} is available as
\textbf{Sophea-Nano-Policy-LIBERO-Greek-v1} at
\url{https://huggingface.co/KIEFERSA/Sophea-Nano-Policy-LIBERO-Greek-v1}: the bilingual
recipe at the seed whose Greek evaluated best of three ($80.2\%$ against a three-seed
mean of $61.3\%$ and a range of $48.6$--$80.2$ on the ten-goal suite), as a Hugging Face
safetensors export
that loads directly into the Cosmos3 policy server, with a model card that reports the
wrong-instruction control alongside the headline numbers and the commands to reproduce the
three-way evaluation. A reader should treat that checkpoint as the maximum of a
three-seed draw rather than the recipe's expectation; the expectation is the mean, and
Section~\ref{sec:study2} gives the spread. Scripts, dose-ladder configurations, training
configurations and the evaluation harness accompany the paper. We are precise about what
that reproduces: the pipeline and every evaluation, but not our exact numbers, because a
reader's own translation of the instructions is a different translator, and
Section~\ref{sec:study2} measures what changing translator costs. Reproducing the numbers
requires our instruction sets, which we will supply on request for research use. Two artifacts are
deliberately not released: the Greek-adapted world model of Section~\ref{sec:study1},
which matches the specific content of a Greek caption on none of the held-out scenes and
should not ship on a coherence number alone, and the translated corpora.

\appendix
\setcounter{topnumber}{3}\setcounter{totalnumber}{4}
\renewcommand{\topfraction}{0.98}\renewcommand{\textfraction}{0.01}\renewcommand{\floatpagefraction}{0.5}
\section{Every inferential comparison}
\label{app:comparisons}

The body makes twenty inferential comparisons and applies no multiplicity
correction. Table~\ref{tab:comparisons} lists them so a reader can apply one. Two
properties of this design matter more than any individual $p$. First, the two families
answer different questions and have different power: \emph{within-policy} contrasts share
a checkpoint, take the goal or task as the sampling unit, and reach $p=0.002$ at ten
clusters; \emph{cross-policy} contrasts take the training run as the sampling unit, and at
three runs per arm the smallest attainable two-sided $p$ is $0.100$. Second, no cross-policy
contrast can therefore survive any correction, and we do not claim one does. Under
Bonferroni at $\alpha=0.05$ over twenty comparisons ($\alpha'=0.0025$) only the
within-policy task-level contrasts survive.

\begin{table*}[t]
\centering
\footnotesize
\setlength{\tabcolsep}{5pt}
\renewcommand{\arraystretch}{1.15}
\begin{tabular}{@{}p{0.34\textwidth}p{0.07\textwidth}p{0.21\textwidth}p{0.20\textwidth}p{0.08\textwidth}@{}}
\toprule
\textbf{Comparison} & \textbf{Family} & \textbf{Test / unit} & \textbf{Result} & \textbf{Survives?} \\
\midrule
Greek vs wrong instruction, single-goal suite & within & paired task-level, $n{=}10$ & $+2.0$, $p=0.055$ & no \\
Greek vs wrong, multi-goal suite & within & paired task-level, $n{=}10$ & $p=0.002$ & yes \\
English vs wrong, multi-goal suite & within & paired task-level, $n{=}10$ & $p=0.002$ & yes \\
Greek level, from-base bilingual & within & task-level bootstrap & $48.6$ CI $[25.0, 72.2]$ & n/a \\
Greek vs wrong, LIBERO-90, bilingual, 3 seeds & within & paired task-level, $n{=}90$ & $+7.1/+6.8/+6.7$, CIs exclude $0$ & yes \\
Greek vs wrong, LIBERO-90, target-only, 3 seeds & within & paired task-level, $n{=}90$ & $-0.3/+0.7/+2.7$, $2$ of $3$ include $0$ & partly \\
Trained vs independent phrasings, single-phrasing arm & within & paired, per seed & $-30.8/-24.8/-14.4$ & yes \\
Trained vs independent phrasings, diverse arm & within & paired, per seed & $-12.4/-11.2/-13.4$ & yes \\
Trained vs held-out paraphrase (two sets) & within & paired, per seed & non-monotone; retracted & no \\
Trained vs paraphrased \emph{English}, syntax only & within & paired, per seed & $-2.8/-6.6/-2.4$ & yes \\
Trained vs paraphrased \emph{English}, nouns substituted & within & paired, per seed & $-2.6/-2.2/-9.4$ & yes \\
English vs Greek rewording penalty & within, paired & every-seed ordering, $n{=}3$ pairs & $\max$ EN $9.4 <\min$ EL $14.4$; exact floor here is $0.250$, not $0.100$ & no \\
Failure-mode split (English vs Greek) & within & per-episode goal predicates & $0\%$ vs $82\%$ & yes \\
Single-phrasing vs diverse penalty & cross & exact run-level, $3$v$3$ & $p=0.100$ (floor) & no \\
Bilingual vs target-only, ten goals & cross & exact run-level, $3$v$3$ & $p=0.300$ & no \\
Bilingual vs target-only margin, LIBERO-90 & cross & exact run-level, $3$v$3$ & $p=0.100$ (floor) & no \\
Bilingual vs English-only (no-Greek ablation) & cross & exact run-level, $3$v$3$ & separates & no \\
Frozen vs unfrozen tower & cross & exact run-level, $3$v$3$ & separates & no \\
Warm-started vs from-base & cross & one run per arm & directional only & no \\
English cost of phrasing diversity & cross & task-level CI & $-3.6$ CI $[-6.4, -0.8]$ & no \\
Loss vs Greek success across six policies & cross & correlation, $n{=}6$ & $r=+0.744$, exploratory & no \\
\bottomrule
\end{tabular}
\caption{Every inferential comparison in the paper, its family, its sampling unit, and
whether it survives a Bonferroni threshold at $\alpha'=0.0025$. ``Survives'' is a property
of the design, not of the effect: no run-level contrast at three runs per arm can reach
$0.003$, because $0.100$ is the smallest value that design can produce. The cross-policy
rows should be read as ordering statements with a one-in-ten per-comparison false-positive
rate, and across the five perfect-separation contrasts the probability that at least one
arises by chance is about $41\%$.}
\label{tab:comparisons}
\end{table*}

\section{Per-goal profiles and the Greek instructions}
\label{app:pergoal}

\begin{table*}[t]
\centering
\footnotesize
\setlength{\tabcolsep}{6pt}
\begin{tabular}{@{}clrrrrr@{}}
\toprule
& & \textbf{English} & \multicolumn{3}{c}{\textbf{Greek, by seed}} & \textbf{wrong} \\
\cmidrule(lr){4-6}
\textbf{\#} & \textbf{Goal} & ref.\ run & 42 & 43 & 44 & ref.\ run \\
\midrule
0 & open the middle drawer & 100 & 8 & 8 & 0 & 0 \\
1 & bowl on the stove & 98 & 24 & 60 & 58 & 0 \\
2 & wine bottle on cabinet & 98 & 0 & 94 & 0 & 0 \\
3 & open top drawer, bowl inside & 92 & 24 & 86 & 92 & 0 \\
4 & bowl on top of cabinet & 96 & 94 & 100 & 94 & 44 \\
5 & push plate to front of stove & 98 & 72 & 96 & 96 & 0 \\
6 & cream cheese in the bowl & 94 & 2 & 94 & 36 & 0 \\
7 & turn on the stove & 100 & 90 & 98 & 78 & 0 \\
8 & bowl on the plate & 100 & 98 & 98 & 98 & 0 \\
9 & wine bottle on the rack & 88 & 74 & 68 & 0 & 0 \\
\midrule
& goals $\geq 50\%$ & 10 & 5 & 9 & 6 & 0 \\
\bottomrule
\end{tabular}
\caption{Per-goal success (\%) on the ten-goal suite, 50 trials per cell. The Greek columns
are three seeds of one recipe; English and wrong-instruction are the reference run. Goal~4
carries the wrong-instruction attractor described in Section~\ref{sec:study2}, which is
why its control column is $44$ where every other is $0$.}
\label{tab:pergoal}
\end{table*}

Section~\ref{sec:study2} reports that Greek success is bimodal across goals and that the
split moves with the seed. Table~\ref{tab:pergoal} gives the underlying profile for all
three seeds of the bilingual recipe, together with the English and wrong-instruction
columns of the reference run, and Figure~\ref{fig:strings} gives the ten Greek instructions
so a reader can judge translation adequacy rather than take our word for it.

Two things in these tables constrain explanations we would otherwise be free to offer.
First, the number of goals above $50\%$ is five, nine and six across seeds and only goal~0
stays below it in every seed, so ``five succeed and five fail'' describes one run, not the
recipe. Second, the obvious mechanical explanation for a persistent per-goal failure, that
the translation of that goal is wrong, does not hold for the one goal that persistently
fails: \emph{open the middle drawer of the cabinet} renders literally and unambiguously (Figure~\ref{fig:strings}, row 0), and the independent translator produces the same sentence. Meanwhile the two
instructions where the glossary does introduce a real ambiguity, goals~1 and~5, where \emph{stove} becomes a word that also means \emph{kitchen} rather than \emph{hob}, score $24/60/58$ and $72/96/96$: one weak, one strong. Translation
adequacy therefore does not predict the per-goal profile in either direction here. We
report this because it was the first explanation we reached for and it did not survive the
strings.

\setlength{\bibsep}{1.6pt plus 0.3pt minus 0pt}
{\small
\bibliographystyle{plainnat}
\bibliography{refs}
}

\end{document}